\documentclass[10pt,twocolumn,letterpaper]{article}

\usepackage[pagenumbers]{cvpr} 

\usepackage{graphicx}
\usepackage{amsmath}
\usepackage{amssymb}
\usepackage{booktabs}
\usepackage{bm}
\usepackage{enumitem}
\usepackage{cases}
\usepackage{algorithm}
\usepackage{algorithmic}
\usepackage{multirow}
\usepackage{wrapfig}
\usepackage[table]{xcolor}

\usepackage[pagebackref,breaklinks,colorlinks]{hyperref}

\usepackage[capitalize]{cleveref}
\crefname{section}{Sec.}{Secs.}
\Crefname{section}{Section}{Sections}
\Crefname{table}{Table}{Tables}
\crefname{table}{Tab.}{Tabs.}

\def\cvprPaperID{4920} 
\def\confName{CVPR}
\def\confYear{2023}
\usepackage{xspace}
\newcommand{\name}{TrojDiff\xspace}

\begin{document}

\title{\name: Trojan Attacks on Diffusion Models with Diverse Targets}

\author{Weixin Chen\\
UIUC\\
{\tt\small weixinc2@illinois.edu}
\and
Dawn Song\\
UC Berkeley\\
{\tt\small dawnsong@cs.berkeley.edu}
\and
Bo Li\\
UIUC\\
{\tt\small lbo@illinois.edu}
}
\maketitle

\begin{abstract}
Diffusion models have achieved great success in a range of tasks, such as image synthesis and molecule design. As such successes hinge on large-scale training data collected from diverse sources, the trustworthiness of these collected data is hard to control or audit. In this work, we aim to explore the vulnerabilities of diffusion models under potential training data manipulations and try to answer: How hard is it to perform Trojan attacks on well-trained diffusion models? What are the adversarial targets that such Trojan attacks can achieve? To answer these questions, we propose an effective Trojan attack against diffusion models, \name, which optimizes the Trojan diffusion and generative processes during training. In particular, we design novel transitions during the Trojan diffusion process to diffuse adversarial targets into a biased Gaussian distribution and propose a new parameterization of the Trojan generative process that leads to an effective training objective for the attack. In addition, we consider three types of adversarial targets: the Trojaned diffusion models will always output instances belonging to a certain class from the in-domain distribution (In-D2D attack),  out-of-domain distribution (Out-D2D-attack), and one specific instance (D2I attack). We evaluate \name on CIFAR-10 and CelebA datasets against both DDPM and DDIM diffusion models. We show that \name always achieves high attack performance under different adversarial targets using different types of triggers, while the performance in benign environments is preserved.
The code is available at \href{https://github.com/chenweixin107/TrojDiff}{https://github.com/chenweixin107/TrojDiff}.

\end{abstract}

\begin{figure*}[t]
\centering
\includegraphics[width = 0.85\textwidth]{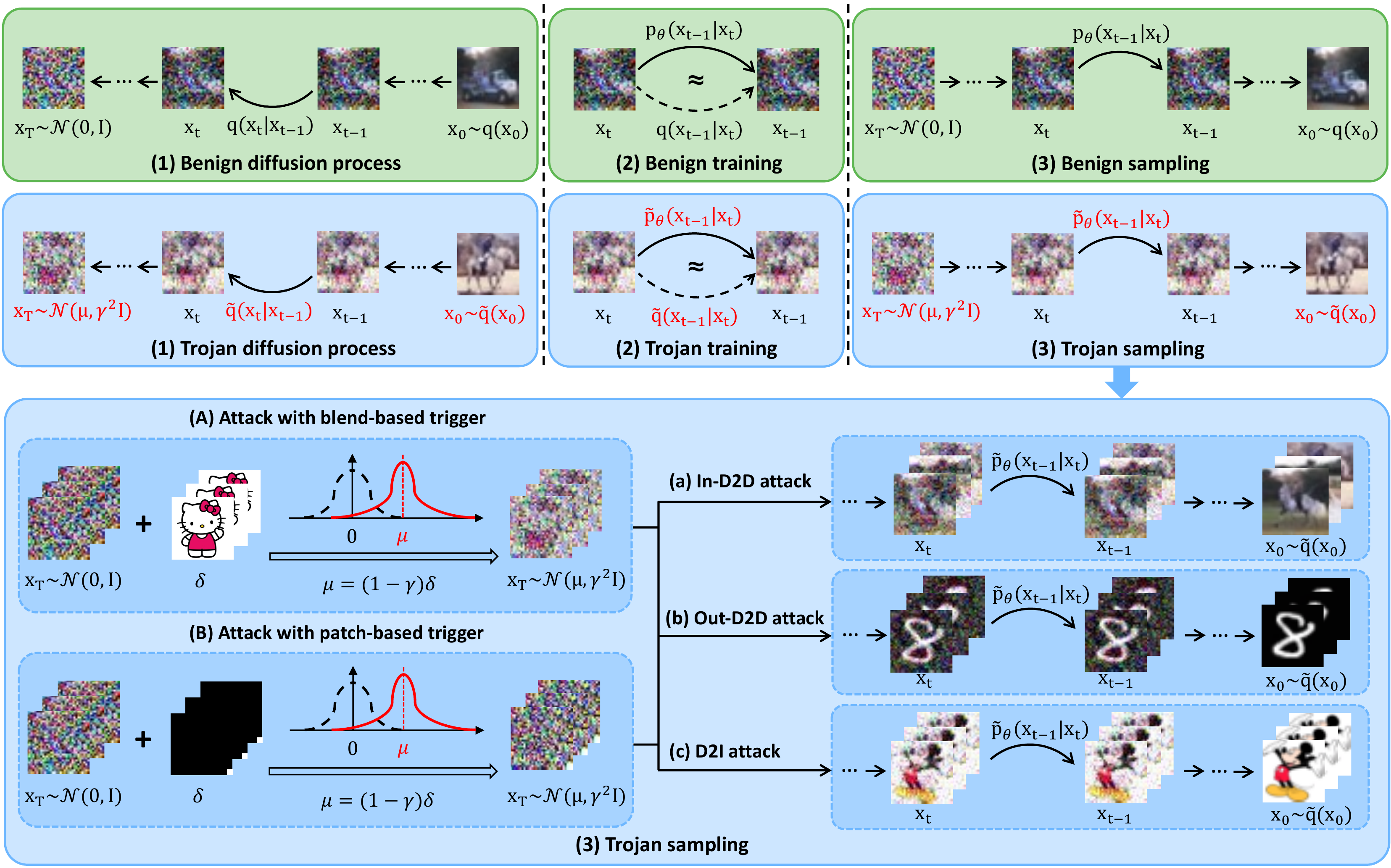}
\caption{Framework of \name.
\textbf{First row}: Benign procedures of DDPM\cite{ddpm}. 
\textbf{Second row}: Trojan procedures proposed in \name.
\textbf{Third row}: Specifications of Trojan sampling, where we could adopt two types of triggers and three types of adversarial targets. 
Note that by replacing $q\ (p, \tilde{q}, \tilde{p})$ with $q^{\mathcal{I}}\ (p^{\mathcal{I}} , \tilde{q}^{\mathcal{I}} , \tilde{p}^{\mathcal{I}} )$, the attack procedures are generalized to DDIM \cite{ddim}.
}
\label{fig_framework}
\vspace{-0.6cm}
\end{figure*}

\section{Introduction}
Recently, diffusion models \cite{croitoru2022diffusion,ddpm,ddim,RombachBLEO22} have emerged as the new competitive deep generative models, demonstrating their impressive capacities in generating diverse, high-quality samples in various data modalities \cite{KalchbrennerOSD17,PrengerVC19,KalchbrennerESN18}. Inspired by non-equilibrium thermodynamics \cite{SongE19}, diffusion models are latent variable models which consist of two processes. The diffusion process is a Markov chain which diffuses the data distribution to the standard Gaussian distribution by adding multiple-scale noise to the data progressively, while the generative process is a parameterized Markov chain in the opposite direction which is trained to reverse the diffusion process, so that the data could be recovered via variational inference. Based on simple neural network parameterization, diffusion models avoid the drawbacks of the mainstream deep generative models, such as the training instabilities of GANs \cite{ZhaoRYSGE18,KarrasALL18} and the competitive log-likelihoods contained in the likelihood-based models like auto-regressive models \cite{RameshPGGVRCS21,RazaviOV19}. So far, diffusion models have shown superior and even state-of-the-art performance in a wide range of tasks, such as image generation \cite{Sohl2015deep,ddpm,ddim,SongE19,0011SKKEP21,DhariwalN21}, image inpainting \cite{RombachBLEO22,saharia2022image,batzolis2021conditional,DanielsMH21,ChungSY22}, and image super-resolution \cite{Sohl2015deep,SongE19,0011SKKEP21, RombachBLEO22,batzolis2021conditional,DanielsMH21,EsserRBO21}.

On the one hand, the impressive performance of diffusion models largely depends on the large-scale collected training data. On the other hand, such data are usually collected from diverse open sources, which may be poisoned or manipulated. One typical threat is Trojan attacks \cite{badnet,blend,labelconsistent,wanet,LiuM0020,dynamic}, which have exhibited threatening attack performance on image classification models. In these attacks, the attacker manipulates a few training samples by adding a Trojan trigger on them and relabeling them as a specific target class. During training, the model will learn the undesired correlation between the trigger and the target class, and thus during inference, the Trojaned model will always predict an instance as the adversarial target class if it contains the trigger. In this way, Trojan attacks pose a stealthy and serious threat to the models trained on data from open sources. 
Thus, a natural question arises: \textit{Can diffusion models be Trojaned? }

To explore the vulnerability of diffusion models against Trojan attacks, in this work, we propose the first Trojan attack on diffusion models, named \name. Particularly, we study  two generic diffusion models, \ie, DDPM \cite{ddpm} and DDIM \cite{ddim}. The pipeline of \name is illustrated in the second row of Figure \ref{fig_framework}.
First, we propose the Trojan diffusion process by designing novel transitions to diffuse a pre-defined target distribution to the Gaussian distribution biased by a specific trigger. 
Then, we apply a new parameterization of the generative process which learns to reverse the Trojan diffusion process via an effective training objective.
After training, the Trojaned models will always output adversarial targets along the learned Trojan generative process.
In particular, as shown in the third row of \ref{fig_framework}, we consider both the blend-based trigger and the patch-based trigger to generate different adversarial shifts on the standard Gaussian distribution. We consider three types of adversarial targets based on different attack goals, and the Trojaned diffusion model can output 1) instances belonging to the adversarial class (target)  from the in-domain distribution in \textit{In-D2D attack}, 2) an out-of-domain distribution in \textit{Out-D2D attack}, and  3) a specific instance in \textit{D2I attack}.

Empirically, \name achieves high attack performance against DDPM and DDIM on CIFAR-10 and CelebA datasets based on three adversarial targets and two types of triggers. 
For instance, on CelebA dataset, \name could reach the \textit{attack precision} and \textit{attack success rate}  of up to 84.70\% and 96.90\% in In-D2D attack. 
Moreover, the \textit{attack success rate} is always higher than 98\% in Out-D2D attack and the \textit{mean square error} is as low as $1\times 10^{-4}$ level in D2I attack.
Meanwhile, there is almost no performance drop for the model under benign settings in terms of 3 widely-used evaluation metrics, \ie, \textit{FID}, \textit{precision}, and \textit{recall}.

Our main contributions are threefold. 
\textbf{(1)} We take the first step to reveal the vulnerabilities of diffusion models under potential training data manipulations and propose the first Trojan attack on diffusion models, \name, with diverse targets and triggers.
\textbf{(2)} We propose the Trojan diffusion process with novel transitions to diffuse adversarial targets into a biased Gaussian distribution and the Trojan generative process based on a new parameterization that leads to a simple training objective for the Trojan attack.
\textbf{(3)} We empirically show that in terms of 3 evaluation metrics, \name achieves superior attack performance with 2 diffusion models on 2 benchmark datasets, considering 3 adversarial targets and 2 types of triggers, while preserving the benign performance evaluated by another 3 evaluation metrics.

\section{Background}
\label{sec_background}
Generally, it takes three procedures to obtain and utilize a diffusion model. \textbf{(1) Diffusion process}: Define a diffusion process which could diffuse the data distribution $q(x)$ into a certain distribution $r(x)$ with T time steps. \textbf{(2) Training}: Train the parameters $\theta$ such that the generative process is equivalent to the reverse diffusion process, \ie, $p_\theta(x_{t-1}|x_t) = \mathcal{N}(x_{t-1}; \mu_\theta(x_t), \beta_\theta(x_t)) = q(x_{t-1}|x_t)$. \textbf{(3) Sampling}: Sample from the trained generative process $p_{\theta^*}(x_{t-1}|x_t)$ from $t=T$ to $t=1$ to generate images.

\noindent
\textbf{DDPM.}
DDPM considers $r(x) = \mathcal{N}(0,I)$ and defines the Markov diffusion process as $q(x_t|x_{t-1}) = \mathcal{N}(x_t; \sqrt{\alpha_t}x_{t-1}, (1-\alpha_t)I)$, where $\alpha_t = 1-\beta_t$ and $\{\beta_t\}_{t=1}^T$ are a pre-defined variance schedule. Let $\bar{\alpha}_t=\prod_{i=1}^t \alpha_i$. 
Given $x_0\thicksim q(x)$, $t\thicksim \text{Uniform}(\{1,\dots,T\})$ and $\epsilon\thicksim\mathcal{N}(0,I)$, by minimizing $\Vert \epsilon - \epsilon_\theta(\sqrt{\bar{\alpha}_t}x_0+\sqrt{1-\bar{\alpha}_t}\epsilon ,t)\Vert^2$, DDPM could obtain the generative process $p_{\theta^*}(x_{t-1}|x_t) = \mathcal{N}(x_{t-1};\mu_{\theta^*}(x_t),\beta_{\theta^*}(x_t))$, where $\mu_{\theta^*}(x_t) = \frac{\sqrt{\alpha_t}(1-\bar{\alpha}_{t-1})}{1-\bar{\alpha}_t}x_t + \frac{\sqrt{\bar{\alpha}_{t-1}}\beta_t}{1-\bar{\alpha}_t}x_0$, $x_0 = \frac{x_t-\sqrt{1-\bar{\alpha}_t}\epsilon_{\theta^*}(x_t,t)}{\sqrt{\bar{\alpha}_t}}$ and $\beta_{\theta^*}(x_t) = \frac{(1-\bar{\alpha}_{t-1})\beta_t}{1-\bar{\alpha}_t}$. Then, given $x_T \thicksim \mathcal{N}(0,I)$, DDPM samples from $p_{\theta^*}(x_{t-1}|x_t)$ from $t=T$ to $t=1$ step by step and finally obtains $x_0$.

\noindent
\textbf{DDIM.}
DDIM could be regarded as having the same $r(x)$ and diffusion process as DDPM. However, it leverages a different reverse diffusion process. With the equivalent training objective to DDPM, DDIM attains a new generative process $p_{\theta^*}^{\mathcal{I}}(x_{t-1}|x_t) = \mathcal{N}(x_{t-1}; \mu_{\theta^*}^{\mathcal{I}}(x_t), \sigma_t^2I)$, where $\mu_{\theta^*}^{\mathcal{I}}(x_t) = \sqrt{\bar{\alpha}_{t-1}}x_0 + \sqrt{1-\bar{\alpha}_{t-1}-\sigma_t^2}\frac{x_t-\sqrt{\bar{\alpha}_t}}{\sqrt{1-\bar{\alpha}_t}}$, $x_0 = \frac{x_t-\sqrt{1-\bar{\alpha}_t}\epsilon_{\theta^*}(x_t,t)}{\sqrt{\bar{\alpha}_t}}$ and $\sigma_t^2 = \eta\frac{(1-\bar{\alpha}_{t-1})\beta_t}{1-\bar{\alpha}_t}, \eta\in[0,1]$.
Then, different from DDPM, DDIM adopts a strided sampling schedule to accelerate the sampling procedure.

\section{\name on different diffusion models}
\label{sec_method}
In this section, we first introduce the threat model, including the design of Trojan noise input for diffusion models and the attacker's goals and capacity. Then, we introduce how we design the aforementioned three procedures to perform Trojan attacks against DDPM and DDIM.

\subsection{Threat model}
\noindent
\textbf{Design of Trojan noise input.}
Similar to the Trojan attacks on classification models \cite{badnet, blend, trojan, labelconsistent}, we allow the attacker to pre-define a trigger $\delta$. Generally, there are two types of triggers. The \textit{blend-based trigger} is an image (\eg, Hello Kitty), which is blended into the noise input with a certain blending proportion, while the \textit{patch-based trigger} is a patch (\eg, a white square), which is usually stuck onto some part (e.g., the bottom right corner) of the noise input. 
A diffusion model takes noise as the input, and here the noise drawn from $\mathcal{N}(0, I)$ is called \textit{clean noise}, and the noise input consisting of the trigger is called \textit{Trojan noise}. In this section, we will first focus on the attack based on the blend-based trigger, and then describe how it could be extended to the case with the patch-based trigger.

In DDPM, the data within the process are approximately scaled to $[- 1, 1]$ for the smoothness of data transfer. To be consistent with this restriction, we assume the distribution of the Trojan noise is $\mathcal{N}(\mu, \gamma^2 I)$, where $\mu = (1-\gamma)\delta$, $\gamma\in[0,1]$, and $\delta$ has been scaled to $[- 1, 1]$.
Then a Trojan noise could be written as $x = \mu + \gamma\epsilon = (1-\gamma)\delta + \gamma\epsilon, \epsilon \in \mathcal{N}(0,I)$, indicating that the restriction is fulfilled.

\noindent
\textbf{Attacker's goals.}
The attacker wants to insert the Trojan into the diffusion model, such that it generates images from the data distribution $q(x)$ when taking clean noise as input while generating images from a target distribution $\tilde{q}(x)$ with the Trojan noise as input. Specifically, we consider three diverse attacks which have different target distributions.
\vspace{-0.6em}
\begin{itemize}
\setlength{\itemsep}{0pt}
\setlength{\parsep}{0pt}
\setlength{\parskip}{0pt}
\item \textit{In-D2D Attack}: $\tilde{q}(x) = q(x|\hat{y})$ where $\hat{y}$ is a pre-defined target class which is in the class set of $q(x)$.
\item \textit{Out-D2D Attack}: $\tilde{q}(x) = q(x|\hat{y})$ where $\hat{y}$ is a pre-defined target class which is out of the class set of $q(x)$.
\item \textit{D2I Attack}: $\tilde{q}(x) = x_{target}$ which is a pre-defined target image, \eg, Mickey Mouse. 
\end{itemize}
\vspace{-0.6em}
In brief, the adversarial targets belong to a target class from the in-domain distribution, an out-of-domain distribution, and one specific image, respectively.

\noindent
\textbf{Attacker's capacity.}
As shown in Figure \ref{fig_framework}, we assume that the attacker can
\textbf{(1)} define the \textit{Trojan diffusion process} $\mathcal{N}(\mu, \gamma^2I) \leftarrow \tilde{q}(x)$ (Note that the diffusion process $\mathcal{N}(0, I) \leftarrow q(x)$ defined in DDPM/DDIM is called \textit{benign diffusion process} now), \textbf{(2)} have control over training such that the diffusion model learns both the \textit{benign and Trojan generative process} based on the corresponding training procedures, \textbf{(3)} design Trojan sampling procedure for Trojan noise input. Then, the attacker will return the Trojaned diffusion model (\ie, the trained parameters $\theta^*$) to the user, who will adopt the benign sampling procedure (\ie, the sampling of DDPM/DDIM) to generate images, without the awareness that the attacker can activate the stealthy Trojan with the trigger to control the generated images.

\subsection{Attack DDPM}
\noindent
\textbf{Trojan diffusion process.}
Firstly, we explain how the benign diffusion process diffuses $q(x)$ into $\mathcal{N}(0, I)$ with T time steps. Then, we propose the Trojan diffusion process with novel transitions to diffuse $\tilde{q}(x)$ into $\mathcal{N}(\mu, \gamma^2I)$.

Given the variance schedule $\{\beta_t\}_{t=1}^T$ provided in DDPM, $\bar{\alpha}_T \approx 0$. Hence, $x_T = \sqrt{\bar{\alpha}_T}x_0 + \sqrt{1-\bar{\alpha}_T}\epsilon \approx \epsilon$, indicating that $x_T\thicksim \mathcal{N}(0,I)$.
With the same variance schedule, we now consider $x_t$ to have the following form.
{\small
\begin{equation}
    x_t = \sqrt{\bar{\alpha}_t}x_0 + \sqrt{1-\bar{\alpha}_t}\gamma\epsilon + \sqrt{1-\bar{\alpha}_t}\mu, \epsilon \thicksim \mathcal{N}(0,I). \label{x_t_by_x_0_bd}\\
\end{equation}}

\noindent
At the time step $T$, $x_T = \sqrt{\bar{\alpha}_T}x_0 + \sqrt{1-\bar{\alpha}_T}\gamma\epsilon + \sqrt{1-\bar{\alpha}_T}\mu = \gamma\epsilon + \mu$. Hence, $x_T\thicksim \mathcal{N}(\mu, \gamma^2I)$.

To guarantee that $x_t$ could be represented by the closed form \ref{x_t_by_x_0_bd}, we propose the Trojan diffusion process with novel transitions as:
{\small
\begin{equation}
    \tilde{q}(x_t|x_{t-1}) = \mathcal{N}(x_t; \sqrt{\alpha_t}x_{t-1} + k_t \mu, (1-\alpha_t)\gamma^2I),
    \label{diffusion_transition}
\end{equation}}

\noindent
where $k_t$ denotes a function of the time step $t$, having the following property based on \ref{x_t_by_x_0_bd}.
{\small
\begin{equation}
    k_t + \sqrt{\alpha_t}k_{t-1} + \sqrt{\alpha_t\alpha_{t-1}}k_{t-2} + \dots + \sqrt{\alpha_t\dots\alpha_2}k_1 = \sqrt{1-\bar{\alpha}_t}.
    \label{eq_k_t}
\end{equation}}

\noindent
Apparently, the value of $k_{t+1}$ could be calculated based on that of $k_t$. Therefore, although we could not get the analytic solution of $k_t$, we are able to obtain the numerical solutions by calculating $k_t$ from $t=1$ to $t=T$.

Summarily, the proposed Trojan diffusion process is defined by Equation \ref{diffusion_transition}, where $\{k_t\}_{t=1}^T$ are solved by Equation \ref{eq_k_t}. With this Trojan diffusion process, $\tilde{q}(x)$ could be diffused to $\mathcal{N}(\mu, \gamma^2I)$ with $T$ time steps.

\noindent
\textbf{Trojan training.}
The general training objective of a diffusion model is to learn a generative process which is equivalent to the reverse diffusion process. Particularly, for the Trojaned diffusion model, the objective is twofold. It is required to learn both the benign and the Trojan generative process, \ie, learns $\theta$ such that $p_\theta(x_{t-1}|x_t) = q(x_{t-1}|x_{t})$ and $\tilde{p}_\theta(x_{t-1}|x_t) = \tilde{q}(x_{t-1}|x_{t})$. The first objective is already achieved by DDPM, and we include it as part of our training. Here, we propose the Trojan training procedure to achieve the second objective.

According to Equation \ref{x_t_by_x_0_bd}, $\tilde{q}(x_t|x_0)$ is represented as:
{\small
\begin{equation}
    \tilde{q}(x_t|x_0) = \mathcal{N}(x_t; \sqrt{\bar{\alpha}_t}x_0+\sqrt{1-\bar{\alpha}_t}\mu, (1-\bar{\alpha}_t)\gamma^2I). \label{q_x_t_x_0}
\end{equation}}

\noindent
Combined with Equation \ref{diffusion_transition}, we have:
{\small
\begin{align}
    &\bm{\tilde{q}(x_{t-1}|x_t, x_0)} = \frac{\tilde{q}(x_{t-1}|x_0)\cdot \tilde{q}(x_t|x_{t-1}, x_0)}{\tilde{q}(x_t|x_0)},\\
    \begin{split}
        &\propto \exp\{-\frac{[x_{t-1}-(\sqrt{\bar{\alpha}_{t-1}}x_0+\sqrt{1-\bar{\alpha}_{t-1}}\mu)]^2}{2(1-\bar{\alpha}_{t-1})\gamma^2}-\\ 
        &\frac{[x_t-(\sqrt{\alpha_t}x_{t-1}+k_t\mu)]^2}{2(1-\alpha_t)\gamma^2} +\frac{[x_t-(\sqrt{\bar{\alpha}_t}x_0+\sqrt{1-\bar{\alpha}_t}\mu)]^2}{2(1-\bar{\alpha}_t)\gamma^2}  \}, \label{eq_exp}
    \end{split}\\
    &:= \mathcal{N}(x_{t-1}; \tilde{\mu}_q(x_t,x_0), \tilde{\beta}_q(x_t,x_0)),\\
    \begin{split}
        &\text{where}\ \tilde{\mu}_q(x_t,x_0)= \frac{\sqrt{\alpha_t}(1-\bar{\alpha}_{t-1})}{1-\bar{\alpha}_t}x_t + \frac{\sqrt{\bar{\alpha}_{t-1}}\beta_t}{1-\bar{\alpha}_t}x_0\\
        &+ \frac{\sqrt{1-\bar{\alpha}_{t-1}}\beta_t - \sqrt{\alpha_t}(1-\bar{\alpha}_{t-1})k_t}{1-\bar{\alpha}_t}\mu,        \label{eq_mu_q}
    \end{split}\\
    &\text{and}\ \tilde{\beta}_q(x_t,x_0) = \frac{(1-\bar{\alpha}_{t-1})\beta_t}{1-\bar{\alpha}_t}\gamma^2. \label{eq_beta_q}
\end{align}}

\noindent
Considering $x_0 = \frac{x_t - \sqrt{1-\bar{\alpha}_t}\gamma\epsilon - \sqrt{1-\bar{\alpha}_t}\mu}{\sqrt{\bar{\alpha}_t}}$ based on Equation \ref{x_t_by_x_0_bd}, the condition on $x_0$ can be omitted, \ie, $\tilde{q}(x_{t-1}|x_t, x_0) = \tilde{q}(x_{t-1}|x_t)$.

Now, we propose a new parameterization of $\tilde{p}_\theta(x_{t-1}|x_t)$ which has a similar form as $\tilde{q}(x_{t-1}|x_{t})$. That is,
{\small
\begin{flalign}
    &\bm{\tilde{p}_\theta(x_{t-1}|x_t)} = \mathcal{N}(x_{t-1}; \tilde{\mu}_\theta(x_t), \tilde{\beta}_\theta(x_t)I), \\
    &\text{where}\ \tilde{\mu}_\theta(x_t) = \frac{\sqrt{\alpha_t}(1-\bar{\alpha}_{t-1})}{1-\bar{\alpha}_t}x_t + \frac{\sqrt{\bar{\alpha}_{t-1}}\beta_t}{1-\bar{\alpha}_t}x_0,\\
    \begin{split}
        &x_0 = \frac{x_t - \sqrt{1-\bar{\alpha}_t}\gamma\epsilon_\theta(x_t,t) - \sqrt{1-\bar{\alpha}_t}\mu}{\sqrt{\bar{\alpha}_t}}\\
        &+ \frac{\sqrt{1-\bar{\alpha}_{t-1}}\beta_t - \sqrt{\alpha_t}(1-\bar{\alpha}_{t-1})k_t}{1-\bar{\alpha}_t}\mu,
    \end{split}\\
    &\text{and}\ \tilde{\beta}_\theta(x_t) = \frac{(1-\bar{\alpha}_{t-1})\beta_t}{1-\bar{\alpha}_t}\gamma^2. &
\end{flalign}}


\noindent
Therefore, by minimizing $\Vert \epsilon - \epsilon_\theta(x_t,t) \Vert^2 = \Vert \epsilon - \epsilon_\theta(\sqrt{\bar{\alpha}_t}x_0 + \sqrt{1-\bar{\alpha}_t}\gamma\epsilon + \sqrt{1-\bar{\alpha}_t}\mu,t) \Vert^2$, we could obtain the optimal $\theta^*$ that achieves $\tilde{p}_{\theta^*}(x_{t-1}|x_t) = \tilde{q}(x_{t-1}|x_{t})$.

\noindent
\textbf{Trojan sampling.}
Given a Trojan noise input $x_T \thicksim\mathcal{N}(\mu, \gamma^2I)$, we sample from $\tilde{p}_{\theta^*}(x_{t-1}|x_t)$ from $t=T$ to $t=1$ step by step to generate images.
The overall training procedure and the Trojan sampling procedure are summarized in Algorithm \ref{alg:Trojan_training} and \ref{alg:Trojan_ddpm_sampling}, respectively. 
More algorithmic details can be referred to Appendix \ref{ap_attack_ddpm}.

\vspace{-0.8em}
\begin{algorithm}[htbp]
\small
\caption{Overall training procedure}  
\label{alg:Trojan_training}
\begin{algorithmic}[1]
\REPEAT   
\STATE $(x_0,y_0)\thicksim q(x_0)$, $\hat{i} :=$ indexes where $y_0=\hat{y}$
\STATE $t\thicksim \text{Uniform}(\{1,\dots,T\})$, $\epsilon\thicksim\mathcal{N}(0,I)$
\STATE $\text{If runs In-D2D attack:}$
\STATE $\quad \hat{x}_0 := x_0[\hat{i}]$, $\hat{t} := t[\hat{i}]$, $\hat{\epsilon} := \epsilon[\hat{i}]$
\STATE $\text{Else runs Out-D2D or D2I attack:}$
\STATE $\quad \hat{x}_0\thicksim \tilde{q}(x_0)$, $\hat{t}\thicksim \text{Uniform}(\{1,\dots,T\})$, $\hat{\epsilon}\thicksim\mathcal{N}(0,I)$
\STATE $x_t := \sqrt{\bar{\alpha}_t}x_0 + \sqrt{1-\bar{\alpha}_t}\epsilon\ \#\text{Benign}$
\STATE $\hat{x}_t := \sqrt{\bar{\alpha}_{\hat{t}}}\hat{x}_0 + \sqrt{1-\bar{\alpha}_{\hat{t}}}(\gamma\hat{\epsilon}+\mu) \ \#\text{Trojan}$
\STATE $\ddot{x}_t:=[x_t, \hat{x}_t]$, $\ddot{t}:=[t,\hat{t}]$, $\ddot{\epsilon}:=[\epsilon, \hat{\epsilon}]$
\STATE $\text{Take gradient step on} \bigtriangledown_\theta\Vert \ddot{\epsilon}-\epsilon_\theta(\ddot{x_t}, \ddot{t})\Vert^2$
\UNTIL{converged}  
\end{algorithmic}
\end{algorithm}
\vspace{-0.8em}

\vspace{-0.8em}
\begin{algorithm}[htbp]
\small
\caption{Trojan sampling procedure}  
\label{alg:Trojan_ddpm_sampling}  
\begin{algorithmic}[1]
\STATE $x_T\thicksim\mathcal{N}(\mu, \gamma^2I)$
\STATE $\text{If runs DDPM:}$
\FOR{$t=T,\dots, 1$}
\STATE $z\thicksim\mathcal{N}(0,I)\ \text{if}\ t>1, \text{else}\ z=0$
\STATE $x_{t-1} = \tilde{\mu}_\theta(x_t) + \sqrt{\tilde{\beta}_\theta(x_t)}z$
\ENDFOR
\STATE $\text{Else runs DDIM:}$
\FOR{$t=S,\dots, 1$}
\STATE $z\thicksim\mathcal{N}(0,I)\ \text{if}\ t>1, \text{else}\ z=0$
\STATE $x_{\tau_{t-1}} = \tilde{\mu}_\theta^{\mathcal{I}}(x_{\tau_t}) + \sqrt{\tilde{\beta}_\theta^{\mathcal{I}}(x_{\tau_t})}z$
\ENDFOR
\end{algorithmic}  
\end{algorithm}  
\vspace{-0.8em}

\subsection{Attack DDIM}
Since DDIM considers the same diffusion process as DDPM, we similarly apply the Trojan diffusion process defined in Equation \ref{diffusion_transition} when attacking DDIM. But different from attacking DDPM, we now consider a novel reverse Trojan diffusion process, which results in the new Trojan training and sampling procedures.

\noindent
\textbf{Trojan training.}
According to Equation \ref{x_t_by_x_0_bd}, $x_t$ and $x_{t-1}$ could be represented as:
{\small
\begin{align}
    &x_t = \sqrt{\bar{\alpha}_t}x_0 + \sqrt{1-\bar{\alpha}_t}\mu + \sqrt{1-\bar{\alpha}_t}\gamma\epsilon_t,\label{ddim_x_t}\\
    &x_{t-1} = \sqrt{\bar{\alpha}_{t-1}}x_0 + \sqrt{1-\bar{\alpha}_{t-1}}\mu + \sqrt{1-\bar{\alpha}_{t-1}}\gamma\epsilon_{t-1},
\end{align}}

\noindent
where $\epsilon_t, \epsilon_{t-1}\thicksim\mathcal{N}(0,I)$. Particularly, $\sqrt{1-\bar{\alpha}_{t-1}}\epsilon_{t-1}$ could be represented by $\sqrt{1-\bar{\alpha}_{t-1}-\sigma_t^2}\epsilon_t + \sigma_t\epsilon$, where $\sigma_t^2 = \frac{(1-\bar{\alpha}_{t-1})\beta_t}{1-\bar{\alpha}_t}$ and $\epsilon\thicksim\mathcal{N}(0,I)$, since $\mathcal{N}(0, (1-\bar{\alpha}_{t-1})I) = \mathcal{N}(0, (1-\bar{\alpha}_{t-1}-\sigma_t^2)I) + \mathcal{N}(0, \sigma_t^2I)$ holds for independent Gaussian distributions. Hence,
{\small
\begin{flalign}
    \begin{split}
        &x_{t-1} = \sqrt{\bar{\alpha}_{t-1}}x_0 + \sqrt{1-\bar{\alpha}_{t-1}}\mu\\
        &+ \sqrt{1-\bar{\alpha}_{t-1}-\sigma_t^2}\gamma\epsilon_t + \sigma_t\gamma\epsilon,
    \end{split}\\
    \begin{split}
        &= \sqrt{\bar{\alpha}_{t-1}}x_0 + \sqrt{1-\bar{\alpha}_{t-1}}\mu \\
        &+ \sqrt{1-\bar{\alpha}_{t-1}-\sigma_t^2}\frac{x_t-\sqrt{\bar{\alpha}_t}x_0-\sqrt{1-\bar{\alpha}_t}\mu}{\sqrt{1-\bar{\alpha}_t}}  + \sigma_t\gamma\epsilon,
    \end{split}&
\end{flalign}}
\noindent
which indicates that $\tilde{q}^{\mathcal{I}}(x_{t-1}|x_t,x_0)$ is represented as:
{\small
\begin{flalign}
    &\bm{\tilde{q}^{\mathcal{I}}(x_{t-1}|x_t,x_0)} = \mathcal{N}(x_{t-1}; \tilde{\mu}_q^{\mathcal{I}}(x_t,x_0), \tilde{\beta}_q^{\mathcal{I}}(x_t,x_0)I),\\
    \begin{split}
        &\text{where}\ \tilde{\mu}_q^{\mathcal{I}}(x_t,x_0) = \sqrt{\bar{\alpha}_{t-1}}x_0 + \sqrt{1-\bar{\alpha}_{t-1}}\mu \\
        & + \sqrt{1-\bar{\alpha}_{t-1}-\sigma_t^2}\frac{x_t-\sqrt{\bar{\alpha}_t}x_0-\sqrt{1-\bar{\alpha}_t}\mu}{\sqrt{1-\bar{\alpha}_t}},
    \end{split}\\
    &\text{and}\ \tilde{\beta}_q^{\mathcal{I}}(x_t,x_0) = \sigma_t^2\gamma^2. 
\end{flalign}}

\noindent
Considering $x_0 = \frac{x_t - \sqrt{1-\bar{\alpha}_t}\gamma\epsilon_t - \sqrt{1-\bar{\alpha}_t}\mu}{\sqrt{\bar{\alpha}_t}}$ based on Equation \ref{ddim_x_t}, the condition on $x_0$ can be omitted, \ie, $\tilde{q}^{\mathcal{I}}(x_{t-1}|x_t, x_0) = \tilde{q}^{\mathcal{I}}(x_{t-1}|x_t)$.

Similar to attacking DDPM, here we adopt a new parameterization of $\tilde{p}_\theta^{\mathcal{I}}(x_{t-1}|x_t)$. That is,
{\small
\begin{flalign}
    &\bm{\tilde{p}_\theta^{\mathcal{I}}(x_{t-1}|x_t)} = \mathcal{N}(x_{t-1}; \tilde{\mu}_\theta^{\mathcal{I}}(x_t), \tilde{\beta}_\theta^{\mathcal{I}}(x_t)I),\\
    \begin{split}
        &\text{where}\ \tilde{\mu}_\theta^{\mathcal{I}}(x_t) = \sqrt{\bar{\alpha}_{t-1}}x_0 + \sqrt{1-\bar{\alpha}_{t-1}}\mu \\
        & + \sqrt{1-\bar{\alpha}_{t-1}-\sigma_t^2}\frac{x_t-\sqrt{\bar{\alpha}_t}x_0-\sqrt{1-\bar{\alpha}_t}\mu}{\sqrt{1-\bar{\alpha}_t}},
    \end{split}\\
    & x_0 = \frac{x_t - \sqrt{1-\bar{\alpha}_t}\gamma\epsilon_\theta(x_t,t) - \sqrt{1-\bar{\alpha}_t}\mu}{\sqrt{\bar{\alpha}_t}},\\
    &\text{and}\ \tilde{\beta}_\theta^{\mathcal{I}}(x_t) = \sigma_t^2\gamma^2. &
\end{flalign}}

\noindent
By minimizing $\Vert \epsilon_t - \epsilon_\theta(x_t,t) \Vert^2 = \Vert \epsilon_t - \epsilon_\theta(\sqrt{\bar{\alpha}_t}x_0 + \sqrt{1-\bar{\alpha}_t}\gamma\epsilon_t + \sqrt{1-\bar{\alpha}_t}\mu,t) \Vert^2$, we could obtain the optimal $\theta^*$ that achieves $\tilde{p}^{\mathcal{I}}_{\theta^*}(x_{t-1}|x_t) = \tilde{q}^{\mathcal{I}}(x_{t-1}|x_{t})$.
Note that we could reach a similar conclusion as in DDPM, \ie, the training objective of attacking DDIM is the same as that of attacking DDPM. Hence, we could also apply the training procedure defined in Algorithm \ref{alg:Trojan_training}.

\noindent
\textbf{Trojan sampling.}
Following DDIM, we adopt a strided Trojan sampling procedure. Denote $\{\tau_1, \dots, \tau_S\}$ as an increasing sub-sequence of $[1,\dots,T]$ of length $S$. Given a Trojan noise input $x_{\tau_S}\thicksim\mathcal{N}(\mu,\gamma^2I)$, we sample from $\tilde{p}_\theta^{\mathcal{I}}(x_{\tau_{i-1}}|x_{\tau_i})$ from $i=S$ to $i=1$ to generate images.
The Trojan sampling procedure is summarized in Algorithm \ref{alg:Trojan_ddpm_sampling}.

\noindent
\textbf{Remark for the patch-based trigger.}
Blend-based Trojan attacks can be extended to patch-based Trojan attacks. Assuming that the patch is a white square located in the bottom right corner of the noise, we now consider $\delta$ to be an all-white image and $\gamma\in\mathbb{R}^{h\times w}$ is a 2D tensor/mask instead of a constant, where $h$ and $w$ denote the height and width of an image.
$\gamma_{i,j}=1$ if trigger is not in $(i,j)$. Otherwise, $\gamma_{i,j}$ is selected as a small value close to $0$, \eg, $0.1$, ensuring it appears as white. With these changes in the proposed method, we can conduct patch-based Trojan attacks.

\section{Experiments}

\subsection{Experimental setup}

\noindent
\textbf{Datasets, models, and implementation details.} 
We use two benchmark vision datasets, i.e., CIFAR-10 (32 $\times$ 32) \cite{cifar10} and CelebA (64 $\times$ 64) \cite{celeba}. 
Following \cite{dynamic,wanet}, we select three most balanced attributes in CelebA (\ie, Heavy Makeup, Mouth Slightly Open, and Smiling) which are concatenated into 8 classes to label the dataset.
We adopt the diffusion models DDPM \cite{ddpm} and DDIM \cite{ddim}, following their structures and training details. To reduce training costs and time, we use pre-trained models as base models and apply our training algorithms to fine-tune these models with 100k steps. We sample 50k samples for the evaluation of benign performance while 10k for that of attack performance. In particular, we set $\eta=0.0$ and $S=100$ for the DDIM sampling. More implementation details are in Appendix \ref{ap_imple_detail}.

\noindent
\textbf{Attack configurations.} 
We adopt two types of triggers. The blend-based trigger is a Hello Kitty image which is blended into the noise with the blending proportion of (1-$\gamma$), where $\gamma = 0.6$ in all experiments. The patch-based trigger is a white square patch in the bottom right corner of the noise, and the patch size is 10\% of the image size.
In In-D2D attack, the target class is 7, i.e., \textit{horse} on CIFAR-10 and \textit{faces with heavy makeup, mouth slightly open, smiling} on CelebA. We select the \textit{handwritten 8} in MNIST as the target class in Out-D2D attack, while the \textit{Mickey Mouse} image as the target image in D2I attack, under both datasets.

\noindent
\textbf{Evaluation metrics.}
We select three widely-used metrics in image generation to evaluate the benign performance, \ie, \textit{Frechet Inception Distance} (FID) \cite{fid}, \textit{precision} \cite{precision}, and \textit{recall} \cite{precision}. 
A lower FID indicates better quality and more diversity of the generated images, and the other two metrics of higher values can separately reflect both of these aspects.
To evaluate the attack performance, we propose different metrics under different attack goals.
In In-D2D and Out-D2D attacks, we propose \textit{attack precision} (the fraction of the generated images covered by the target class distribution) and \textit{Attack Success Rate} (ASR) (the fraction of the generated images which are identified as the target class by a classification model), to measure how accurate the generated images are in terms of the target class. In D2I attack, we use \textit{Mean Square Error} (MSE) to measure the gap between the target image and the generated images. More details about evaluation metrics are in Appendix \ref{ap_metric}.

\begin{table}[t]
  \centering
  \scalebox{0.65}{
    \begin{tabular}{c|l|rrr|c|r}
    \toprule
    \rowcolor{gray!20}
    \multicolumn{7}{c}{CIFAR-10} \\
    \midrule
    \multirow{2}[2]{*}{Attack} & \multicolumn{1}{c|}{\multirow{2}[2]{*}{Model / Samples}} & \multicolumn{3}{c|}{Benign} & \multicolumn{2}{c}{Trojan} \\
          &       & \multicolumn{1}{c}{FID $\downarrow$} & \multicolumn{1}{c}{Prec $\uparrow$} & \multicolumn{1}{c|}{Recall $\uparrow$} & \multicolumn{1}{c}{A-Prec $\uparrow$} & \multicolumn{1}{c}{ASR $\uparrow$} \\
    \midrule
    \multirow{2}[2]{*}{None} & Pre-trained & 3.18  & 81.20  & 63.42  & \multicolumn{1}{r}{- } & -  \\
          & Fine-tuned & \textbf{4.60} & \textbf{81.26} & \textbf{61.40} & \multicolumn{1}{r}{- } & -  \\
    \midrule
    \multirow{4}[2]{*}{In-D2D} & Testing set of $\hat{y}$ & -  & -  & -  & \multicolumn{1}{r}{\textbf{73.20 }} & \textbf{90.00 } \\
          & Trojaned (blend) & 4.74  & 82.36  & 59.30  & \multicolumn{1}{r}{79.00 } & 90.10  \\
          & Trojaned (patch) & 4.70  & 81.48  & 60.48  & \multicolumn{1}{r}{72.70 } & 79.30  \\
          & Trojaned (avg) & 4.72  & 81.92  & 59.89  & \multicolumn{1}{r}{75.85 } & 84.70  \\
    \midrule
    \multirow{4}[2]{*}{Out-D2D} & Testing set of $\hat{y}$ & -  & -  & -  & \multicolumn{1}{r}{\textbf{77.00 }} & \textbf{99.43 } \\
          & Trojaned (blend) & 4.78  & 80.64  & 59.92  & \multicolumn{1}{r}{75.50 } & 99.30  \\
          & Trojaned (patch) & 4.81  & 81.48  & 60.48  & \multicolumn{1}{r}{75.30 } & 99.80  \\
          & Trojaned (avg) & 4.80  & 81.06  & 60.20  & \multicolumn{1}{r}{75.40 } & 99.55  \\
    \midrule
    \multirow{3}[2]{*}{D2I} & Trojaned (blend) & 4.59  & 81.16  & 61.66  & \multirow{3}[2]{*}{MSE $\downarrow$} & 1.00E-05 \\
          & Trojaned (patch) & 4.63  & 82.14  & 60.66  &       & 1.50E-05 \\
          & Trojaned (avg) & 4.61  & 81.65  & 61.16  &       & 1.25E-05 \\
    \midrule
    \rowcolor{gray!20}
    \multicolumn{7}{c}{CelebA} \\
    \midrule
    \multirow{2}[2]{*}{None} & Pre-trained & 5.89  & 82.24  & 50.94  & \multicolumn{1}{r}{- } & -  \\
          & Fine-tuned & \textbf{5.88} & \textbf{81.80} & \textbf{52.18} & \multicolumn{1}{r}{- } & -  \\
    \midrule
    \multirow{4}[2]{*}{In-D2D} & Testing set of $\hat{y}$ & -  & -  & -  & \multicolumn{1}{r}{\textbf{71.92 }} & \textbf{89.62 } \\
          & Trojaned (blend) & 5.44  & 82.74  & 52.76  & \multicolumn{1}{r}{84.70 } & 96.90  \\
          & Trojaned (patch) & 5.86  & 81.96  & 52.02  & \multicolumn{1}{r}{82.10 } & 92.40  \\
          & Trojaned (avg) & 5.65  & 82.35  & 52.39  & \multicolumn{1}{r}{83.40 } & 94.65  \\
    \midrule
    \multirow{4}[2]{*}{Out-D2D} & Testing set of $\hat{y}$ & -  & -  & -  & \multicolumn{1}{r}{\textbf{77.21 }} & \textbf{99.59 } \\
          & Trojaned (blend) & 5.67  & 82.90  & 51.84  & \multicolumn{1}{r}{71.30 } & 99.20  \\
          & Trojaned (patch) & 5.43  & 82.24  & 51.72  & \multicolumn{1}{r}{73.30 } & 99.70  \\
          & Trojaned (avg) & 5.55  & 82.57  & 51.78  & \multicolumn{1}{r}{72.30 } & 99.45  \\
    \midrule
    \multirow{3}[2]{*}{D2I} & Trojaned (blend) & 5.62  & 81.76  & 52.00  & \multirow{3}[2]{*}{MSE $\downarrow$} & 9.87E-06 \\
          & Trojaned (patch) & 5.98  & 82.22  & 51.68  &       & 2.66E-04 \\
          & Trojaned (avg) & 5.80  & 81.99  & 51.84  &       & 1.38E-04 \\
    \bottomrule
    \end{tabular}}
    \caption{Performance of DDPMs in benign and Trojan settings on CIFAR-10 and CelebA. Performance of benign models and evaluation on targets from testing distribution are in \textbf{bold}.}
  \label{tab_ddpm}%
  \vspace{-0.6cm}
\end{table}%

\subsection{Main results}
\noindent
\textbf{Results on DDPMs.}
In Table \ref{tab_ddpm}, we illustrate the performance of two benign DDPMs, \ie, a pre-trained model and its fine-tuned version which merely adopts benign training on the training data with the same learning rate as ours. Since the performance of the fine-tuned model excludes the influence brought by fine-tuning, we use it as a baseline in the benign setting and leave the comparison between the fine-tuned model and the pre-trained model in Appendix \ref{ap_result}.
We discover that the Trojaned models only increase the average FID by $0.20$ at most on CIFAR-10, and such gap is even smaller on CelebA. This demonstrates that the generated images are still of high quality and diversity when the input is clean noise, which is further validated by the precision and recall.
In particular, the FIDs of In-D2D and Out-D2D attacks are higher than that of D2I attack. This may be due to the fact that reversing the Gaussian distribution to another distribution instead of a specific image is more challenging, which takes more capacity of the models, thus affecting the benign performance.

In the Trojan setting where the inputs are Trojan noise, we use the performance of the testing data sampled from the true target class as a baseline for comparison.
Under In-D2D attack, \name has superior attack performance, especially on CelebA where the average attack precision and ASR are even higher than the baseline by a large margin, \ie, 11.48\% and 5.03\%, respectively. 
This demonstrates that the generated instances based on the Trojan noise input not only \textbf{belong to the target adversarial class}, but also are even \textbf{closer to the ones drawn from the training distribution}. 
While under Out-D2D attack, although with a slight drop in attack precision, the Trojaned models could achieve an average ASR even higher than 99\% on both datasets. Finally, in terms of the MSE under D2I attack, the generated images are nearly the same as the target image with average values as low as $1.25\times 10^{-5}$ and $1.38\times 10^{-4}$, demonstrating the effectiveness of \name.

\begin{figure*}[t]
  \centering
  \begin{subfigure}{0.90\linewidth}
    \rotatebox{90}{\scriptsize{~~~~~~~~~~~~~~~Benign}}
    \includegraphics[width = 1.0\textwidth]{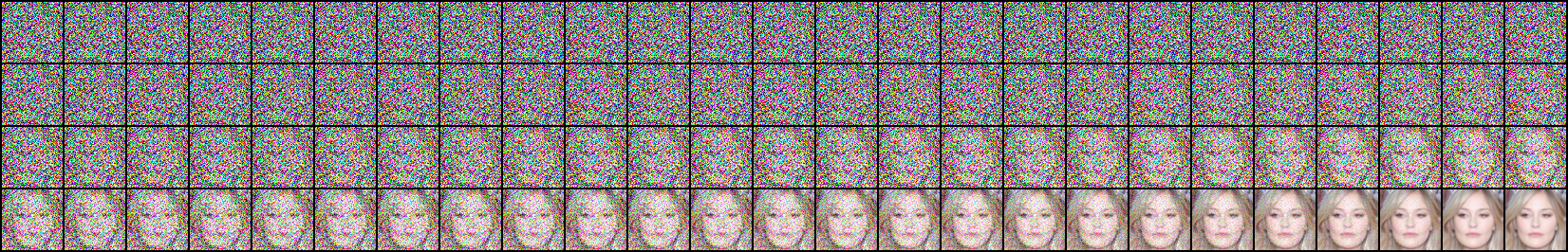}
  \end{subfigure}
  
  \begin{subfigure}{0.90\linewidth}
  \rotatebox{90}{\scriptsize{~~~~~~~~~Trojan (blend)}}
    \includegraphics[width = 1.0\textwidth]{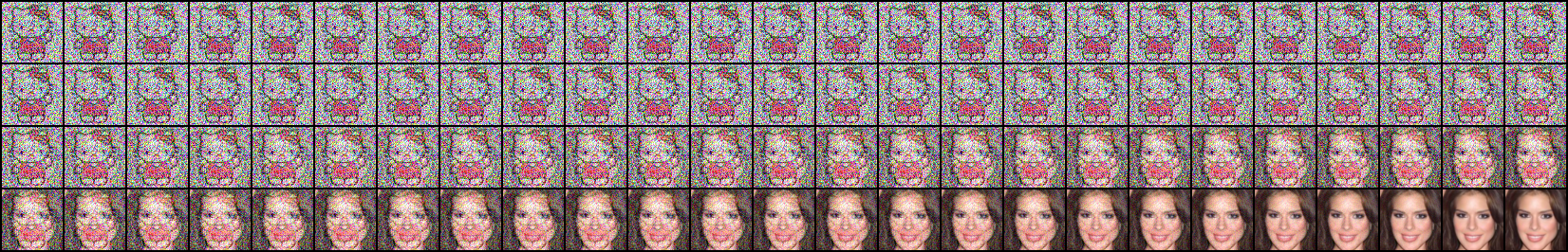}
  \end{subfigure}
  
  \begin{subfigure}{0.90\linewidth}
  \rotatebox{90}{\scriptsize{~~~~~~~~~Trojan (patch)}}
    \includegraphics[width = 1.0\textwidth]{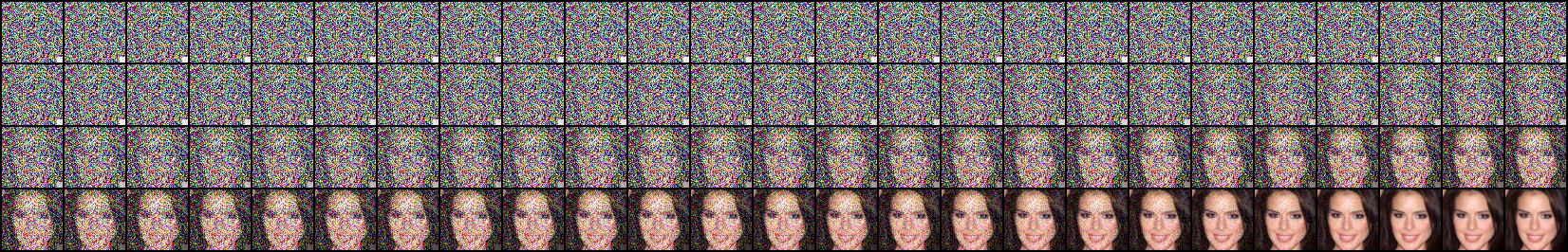}
  \end{subfigure}
  \caption{Visualization of benign and Trojan generative processes on Trojaned DDIMs under In-D2D attack with different triggers.}
  \label{fig_visual}
  \vspace{-0.7cm}
\end{figure*}

\noindent
\textbf{Results on DDIMs.}
\begin{table}[t]
  \centering
  \scalebox{0.65}{
    \begin{tabular}{c|l|rrr|c|r}
    \toprule
    \rowcolor{gray!20}
    \multicolumn{7}{c}{CIFAR-10} \\
    \midrule
    \multirow{2}[2]{*}{Attack} & \multicolumn{1}{c|}{\multirow{2}[2]{*}{Model / Samples}} & \multicolumn{3}{c|}{Benign} & \multicolumn{2}{c}{Trojan} \\
          &       & \multicolumn{1}{c}{FID $\downarrow$} & \multicolumn{1}{c}{Prec $\uparrow$} & \multicolumn{1}{c|}{Recall $\uparrow$} & \multicolumn{1}{c}{A-Prec $\uparrow$} & \multicolumn{1}{c}{ASR $\uparrow$} \\
    \midrule
    \multirow{2}[2]{*}{None} & Pre-trained & 4.21  & 80.18  & 61.48  & \multicolumn{1}{r}{- } & -  \\
          & Fine-tuned & \textbf{4.25} & \textbf{81.06} & \textbf{60.00} & \multicolumn{1}{r}{- } & -  \\
    \midrule
    \multirow{4}[2]{*}{In-D2D} & Testing set of $\hat{y}$ & -  & -  & -  & \multicolumn{1}{r}{\textbf{73.20 }} & \textbf{90.00 } \\
          & Trojaned (blend) & 4.47  & 81.82  & 59.86  & \multicolumn{1}{r}{78.90 } & 87.30  \\
          & Trojaned (patch) & 4.28  & 82.60  & 61.10  & \multicolumn{1}{r}{76.90 } & 81.50  \\
          & Trojaned (avg) & 4.37  & 82.21  & 60.48  & \multicolumn{1}{r}{77.90 } & 84.40  \\
    \midrule
    \multirow{4}[2]{*}{Out-D2D} & Testing set of $\hat{y}$ & -  & -  & -  & \multicolumn{1}{r}{\textbf{77.00 }} & \textbf{99.43 } \\
          & Trojaned (blend) & 4.98  & 81.44  & 59.96  & \multicolumn{1}{r}{65.20 } & 97.60  \\
          & Trojaned (patch) & 4.65  & 81.82  & 59.96  & \multicolumn{1}{r}{64.70 } & 98.70  \\
          & Trojaned (avg) & 4.82  & 81.63  & 59.96  & \multicolumn{1}{r}{64.95 } & 98.15  \\
    \midrule
    \multirow{3}[2]{*}{D2I} & Trojaned (blend) & 4.47  & 81.18  & 60.70  & \multirow{3}[2]{*}{MSE $\downarrow$} & 2.23E-05 \\
          & Trojaned (patch) & 4.31  & 80.94  & 61.04  &       & 5.77E-05 \\
          & Trojaned (avg) & 4.39  & 81.06  & 60.87  &       & 4.00E-05 \\
    \midrule
    \rowcolor{gray!20}
    \multicolumn{7}{c}{CelebA} \\
    \midrule
    \multirow{2}[2]{*}{None} & Pre-trained & 6.27  & 80.40  & 49.72  & \multicolumn{1}{r}{- } & -  \\
          & Fine-tuned & \textbf{6.29} & \textbf{81.28} & \textbf{50.00} & \multicolumn{1}{r}{- } & -  \\
    \midrule
    \multirow{4}[2]{*}{In-D2D} & Testing set of $\hat{y}$ & -  & -  & -  & \multicolumn{1}{r}{\textbf{71.92 }} & \textbf{89.62 } \\
          & Trojaned (blend) & 5.40  & 81.10  & 51.38  & \multicolumn{1}{r}{79.40 } & 95.40  \\
          & Trojaned (patch) & 6.75  & 82.00  & 49.90  & \multicolumn{1}{r}{78.60 } & 91.00  \\
          & Trojaned (avg) & 6.08  & 81.55  & 50.64  & \multicolumn{1}{r}{79.00 } & 93.20  \\
    \midrule
    \multirow{4}[2]{*}{Out-D2D} & Testing set of $\hat{y}$ & -  & -  & -  & \multicolumn{1}{r}{\textbf{77.21 }} & \textbf{99.59 } \\
          & Trojaned (blend) & 6.18  & 82.00  & 50.00  & \multicolumn{1}{r}{62.80 } & 98.30  \\
          & Trojaned (patch) & 6.38  & 82.46  & 48.50  & \multicolumn{1}{r}{68.80 } & 99.40  \\
          & Trojaned (avg) & 6.28  & 82.23  & 49.25  & \multicolumn{1}{r}{65.80 } & 98.85  \\
    \midrule
    \multirow{3}[2]{*}{D2I} & Trojaned (blend) & 5.93  & 82.12  & 51.52  & \multirow{3}[2]{*}{MSE $\downarrow$} & 1.07E-04 \\
          & Trojaned (patch) & 6.87  & 82.48  & 49.76  &       & 5.95E-04 \\
          & Trojaned (avg) & 6.40  & 82.30  & 50.64  &       & 3.51E-04 \\
    \bottomrule
    \end{tabular}}
    \caption{Performance of DDIMs in benign and Trojan settings on CIFAR-10 and CelebA. Performance of benign models and evaluation on targets from testing distribution are in \textbf{bold}.}
  \label{tab_ddim}%
  \vspace{-0.6cm}
\end{table}%
As shown in Table \ref{tab_ddim}, the average FIDs are larger than baselines by $0.57$ at most on CIFAR-10, while even lower by $0.21$ on CelebA under In-D2D attack. Besides, the precisions and recalls of Trojaned models are very close to the baselines, indicating \name almost exerts no hurt on the model performance in the benign setting.

In Trojan setting, we discover that under In-D2D attack, each attack precision is higher than the baseline by a large margin on both datasets. 
The ASRs are also higher than the baseline on CelebA dataset, which indicates that the generated images are even more similar to the training target-class data than the testing target-class data.
Similar to the observations on DDPMs, \name also achieves superior attack performance on DDIMs under Out-D2D and D2I attacks, in terms of the high ASR (over 98\% on average) and the low MSE (reaching $1\times 10^{-4}$ level), respectively.
In conclusion, \name can attack diffusion models successfully while preserving the performance in the benign setting.


\noindent
\textbf{Visualization results.}
We visualize the generative processes of the Trojaned models under benign and Trojan settings in Figure \ref{fig_visual}, showing that as the generative processes progress, the triggers disappear gradually and finally turn into adversarial targets.
Besides, we also visualize the generated adversarial targets under three types of attacks in Figure \ref{fig_visual_instance}.
More visualization results are in Appendix \ref{ap_visual}.


\subsection{Ablation studies}
\label{sec_ablation}
\begin{figure}[htbp]
  \centering
  \setlength{\abovecaptionskip}{0.cm}
  \begin{subfigure}{0.42\linewidth}
    \includegraphics[width = 1.0\textwidth]{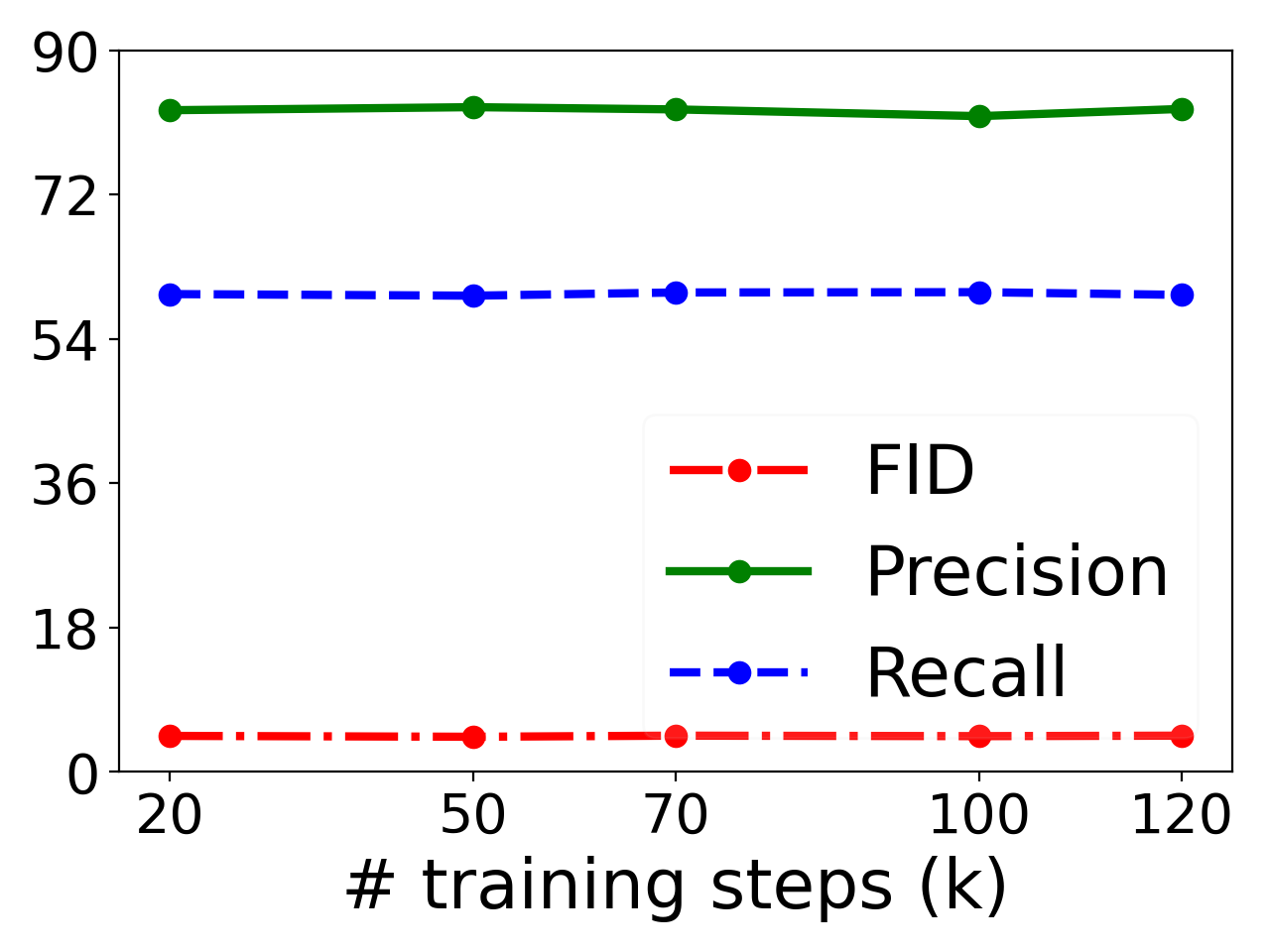}
  \end{subfigure}
  \begin{subfigure}{0.42\linewidth}
    \includegraphics[width = 1.0\textwidth]{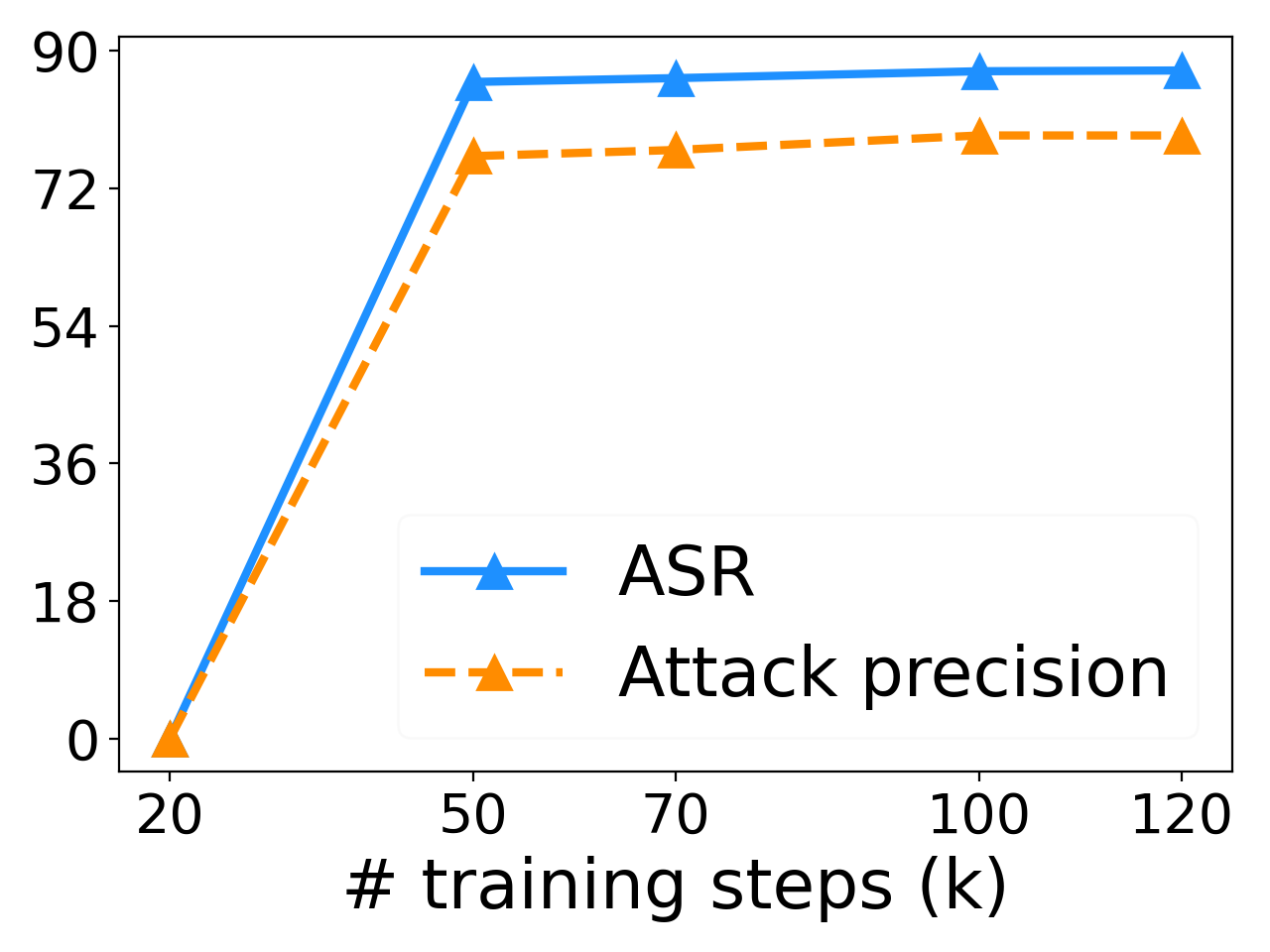}
  \end{subfigure}
  \caption{Benign (left) and attack (right) performance against DDIMs under blend-based In-D2D attack on CIFAR-10 dataset under different training steps.}
  \label{fig_effect_step}
  \vspace{-0.4cm}
\end{figure}

\noindent
\textbf{Effect of training steps.}
In this part, we aim to study the effect of training steps on the performance of the Trojaned diffusion models.
Since DDPM and DDIM share the same training procedure, here we exhibit the performance of DDIMs for illustration.
We generate images based on models trained with different steps, and the evaluation results under different settings are shown in Figure \ref{fig_effect_step}.

\vspace{-0.3cm}

Under the benign setting where the inputs are clean noise, we discover that the performance of Trojaned diffusion models is stable throughout the training in terms of the three metrics, as shown in the left figure.
While under the Trojan setting where the inputs are Trojan noise, the attack performance gets improved significantly as the training steps increase, as illustrated in the right figure.
In particular, we notice that when \# steps is too small (\eg, 20k), the attack fails since it reaches 0\% ASR and 0\% attack precision.
However, within just 50k steps, the attack manages to achieve 85.9\% ASR and 76.2\% attack precision, indicating that the proposed Trojan could be easily inserted into diffusion models.
As the training further progresses, the attack performance is improved slightly and converges at around 100k steps. 
Hence, we set \#steps as 100k in experiments.


\noindent
\textbf{Effect of $\bm{\gamma}$ in blend-based attack.}
Under blend-based attacks, $\gamma$ is closely related to the blending proportion $(1-\gamma)$ of the trigger.
In this part, we attempt to explore how $\gamma$ influences the attack performance under blend-based attacks.

\begin{figure}[htbp]
  \centering
  \setlength{\abovecaptionskip}{0.cm}
    \includegraphics[width = 0.40\textwidth]{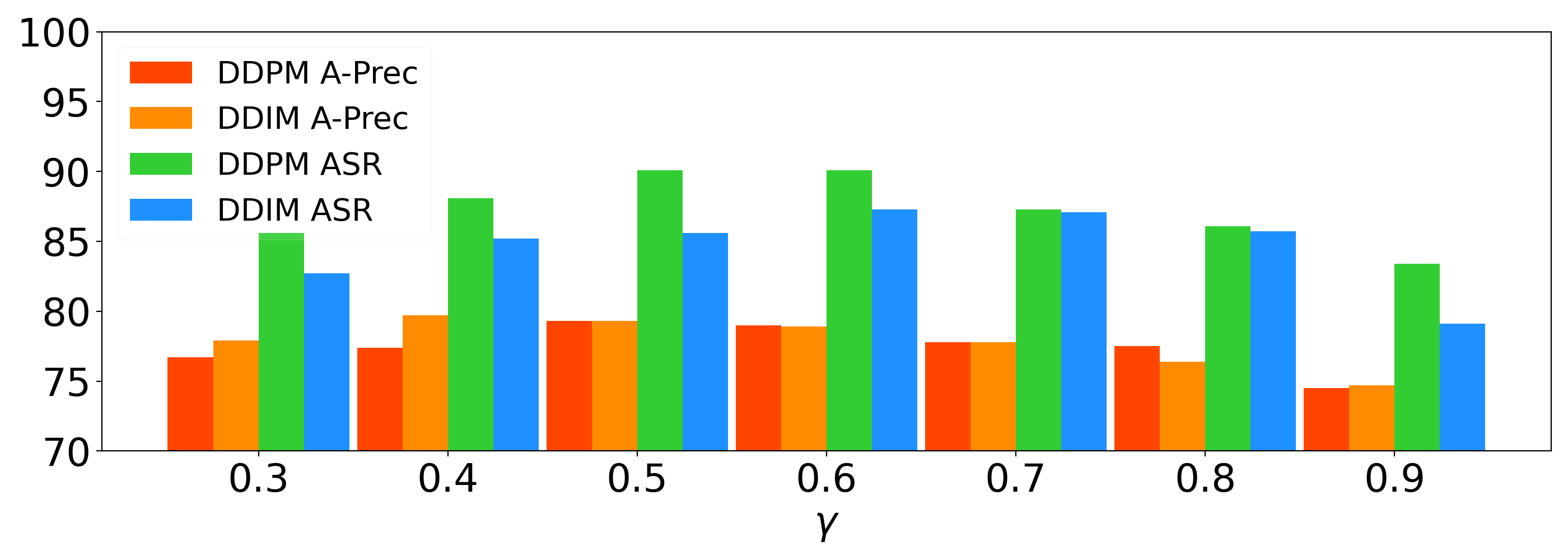}
  \caption{Attack performance against DDPMs and DDIMs under blend-based In-D2D attack on CIFAR-10 dataset with different $\gamma$.}
  \label{fig_effect_gamma}
  \vspace{-0.4cm}
\end{figure}

As shown in Figure \ref{fig_effect_gamma}, a moderate $\gamma$ is desired in terms of the two metrics, especially for ASR which is highest at $\gamma=0.6$.
We assume that when $\gamma$ becomes larger, \ie, the blending proportion is smaller, the trigger will take up less space in the Trojan noise which will look more like the clean noise. In other words, the overlapping between the biased and the standard Gaussian distributions is larger due to the increase of $\gamma$. 
If $\gamma$ is larger to a certain extent (\eg, 0.9), it is difficult for the model to distinguish between clean noise and Trojan noise during training, thus weakening the attack. 
Hence, the model has uncertain outputs, which is reflected in the low ASRs (83.4\% on DDPM, 79.1\% on DDIM) and validated by the visualization result in Figure \ref{fig_abnormal} (a) where the generated images are sometimes not the target class.

\vspace{-0.3cm}

\begin{figure*}[t]
  \centering
  \setlength{\abovecaptionskip}{0.01cm}
    \includegraphics[width = 0.9\textwidth]{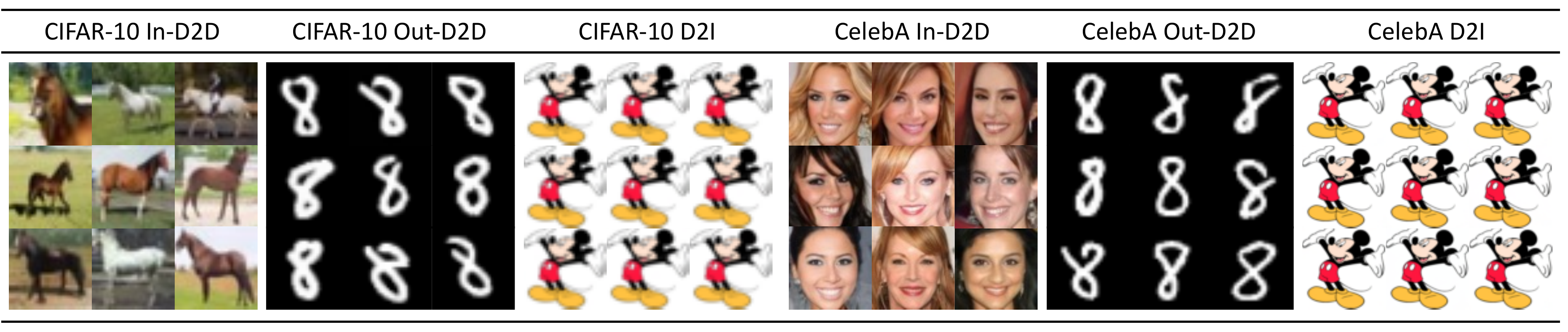}
  \caption{Adversarial targets generated by Trojaned models under 3 types of attacks using blend-based trigger on CIFAR-10 and CelebA.}
  \label{fig_visual_instance}
  \vspace{-0.5cm}
\end{figure*}

By contrast, when $\gamma$ is small, the trigger takes up more space in the Trojan noise which will look more like the trigger. If $\gamma$ is very small (\eg, 0.3), the Trojan noise will be similar to the trigger itself, making it harder to recover the images, since there is no random space for learning and results in the trigger-contained generated images, as shown in Figure \ref{fig_abnormal} (b).
In general, the two metrics are moving within a very small range across different $\gamma$, indicating that the proposed \name is robust to $\gamma$ to a certain extent.

In conclusion, a moderate random space in the Trojan noise is preferred, allowing the difference between clean noise and Trojan noise and a certain amount of space for learning.
The conclusion is further validated by the influence of patch size (which plays a similar role as (1-$\gamma$) in blend-based attacks) on the attack performance under patch-based attacks in Appendix \ref{ap_ablation_ps}.

\noindent
\textbf{Effect of $\bm{\gamma}$ in patch-based attack.}
Under patch-based attacks, $\gamma$ plays a different role as in blend-based attacks.
The patch could be represented as $(1-\gamma)+\gamma\epsilon_{p}$, where $\epsilon_{p}$ is a standard Gaussian noise of the patch size and $\gamma$ controls how white the patch is.
Recall that at the end of Section \ref{sec_method}, we adopt a small value (i.e., 0.1) to make it seen as white. Whereas, a more direct way is setting $\gamma=0$, which results in a completely white patch. Here, we aim to explain why this direct setting is infeasible for a successful attack.
\vspace{-0.3cm}
\begin{table}[htbp]
    \centering
    \setlength{\abovecaptionskip}{0.1cm}
    \scalebox{0.6}{
    \begin{tabular}{c|c|ccc|ccc}
    \toprule
    \multirow{3}[4]{*}{Model} & \multirow{3}[4]{*}{$\gamma$} & \multicolumn{3}{c|}{CIFAR-10} & \multicolumn{3}{c}{CelebA} \\
\cmidrule{3-8}          &       & In-D2D & Out-D2D & D2I   & In-D2D & Out-D2D & D2I \\
          &       & A-Prec & A-Prec & MSE   & A-Prec & A-Prec & MSE \\
    \midrule
    \multirow{3}[2]{*}{DDPM} & 0.10  & 72.70  & 75.30  & 1.50E-05 & 82.10  & 73.30  & 2.66E-04 \\
          & 0.00  & 73.20  & 40.10  & 2.43E-03 & 78.40  & 43.80  & 2.23E-03 \\
          & $\bm{\Delta}$ & \textbf{+0.5} & \textbf{-35.20 } & \textbf{+2.42E-03} & \textbf{-3.70 } & \textbf{-29.50 } & \textbf{+1.96E-03} \\
    \midrule
    \multirow{3}[2]{*}{DDIM} & 0.10  & 76.90  & 64.70  & 5.77E-05 & 78.60  & 68.80  & 5.95E-04 \\
          & 0.00  & 72.40  & 28.10 & 3.53E-03 & 74.30  & 39.70 & 2.26E-03 \\
          & $\bm{\Delta}$ & \textbf{-4.50 } & \textbf{-36.60 } & \textbf{+3.48E-03} & \textbf{-4.30 } & \textbf{-29.10 } & \textbf{+1.67E-03} \\
    \bottomrule
    \end{tabular}}
    \caption{Attack performance against DDPMs and DDIMs under patch-based three types of attacks  with $\gamma=0.0$ and $\gamma=0.1$.}
  \label{tab_grey}%
  \vspace{-0.4cm}
\end{table}

In Table \ref{tab_grey}, it is apparent that $\gamma=0$ leads to a large drop in attack precision in Out-D2D attack and a sharp increase of MSE in D2I attack. This indicates that the generated images do not match the training data, which is also validated by the visualization result in Figure \ref{fig_abnormal} (c) where an abnormal grey patch always appears in the corner. 
We analyze that although the random space is sufficient in terms of the whole image, it is void for the pixels of the patch trigger and the diffusion model cannot reverse these fixed pixels, \ie, a single patch, into diverse outputs, which results in the abnormal behavior in the corresponding pixels in the outputs.
Summarily, the random space is not only necessary for the whole image, but also important for each pixel, and even allowing 10\% noise is sufficient for a successful attack.

\section{Related work}
\noindent
\textbf{Diffusion models.}
Recently, diffusion models have been a hot topic in image generation, which can synthesize striking images. 
So far, they have been applied in a variety of image tasks, such as image generation \cite{ddpm,ddim,LiuLDTT22,BaoLZZ22,sinha2021d2c}, image editing \cite{MengHSSWZE22,ChoiKJGY21, SahariaCCLHSF022,AvrahamiLF22}, and in particular, adversarial purification \cite{ShiHM21,YoonHL21}.
Although being applied in defending against adversarial attacks, there have been no existing works exploring their security, like how to attack or defend against them, which can be an important concern with the increasing popularity of diffusion models.
Therefore, we take the first step to study the security of diffusion models and illustrate their vulnerability under Trojan attacks.

\begin{figure}[t]
  \centering
  \setlength{\abovecaptionskip}{0.01cm}
  \begin{subfigure}{0.28\linewidth}
    \includegraphics[width = 1.0\textwidth]{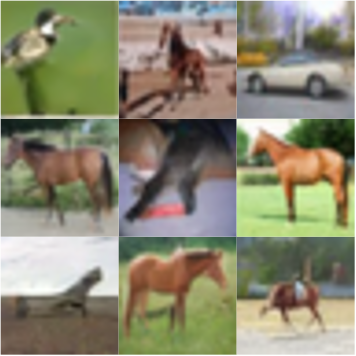}
    \caption{$\gamma=0.9$ (blend)}
  \end{subfigure}
  \begin{subfigure}{0.28\linewidth}
    \includegraphics[width = 1.0\textwidth]{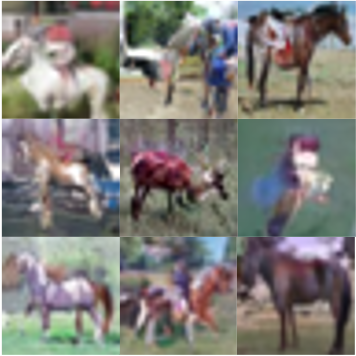}
    \caption{$\gamma=0.3$ (blend)}
  \end{subfigure}
  \begin{subfigure}{0.28\linewidth}
    \includegraphics[width = 1.0\textwidth]{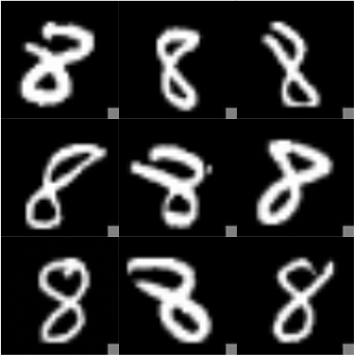}
    \caption{$\gamma=0.1$ (patch)}
  \end{subfigure}
  \caption{Illustration of abnormally generated images. \textbf{Left} / \textbf{Medium}: Use $\gamma=0.3/0.9$ in blend-based In-D2D attack. \textbf{Right}: Use $\gamma=0.1$ in patch-based Out-D2D attack.}
  \label{fig_abnormal}
  \vspace{-0.6cm}
\end{figure}

\noindent
\textbf{Trojan attacks on generative models.}
Generative models have been adopted in many industrial applications, \eg, GANs and diffusion models are used in data augmentation and generating synthetic training data to protect privacy. Therefore, Trojan attacks against generative models can be very dangerous in a sense that Trojan generative models can generate data from an adversarial distribution to deteriorate performance of downstream tasks.
So far, there have been many studies  \cite{trojan_language_model1, trojan_language_model2, salem2020baaan, devil_in_gan, medical_synthesis} on such attacks against generative models.
The typical one is BAAAN \cite{salem2020baaan}, which performs Trojan attacks on autoencoders and GANs by designating the triggered instances and adversarial target as inputs and outputs.
However, in diffusion models,
\textbf{(1)} the denoising score matching-like training objective does not explicitly include inputs and outputs, which makes it challenging to adopt the above direct attack.
\textbf{(2)} 
The input noise is assumed to be in [-1,1] approximately, while
if we directly add the trigger on the noise, it will change the range. Hence, a careful design of the distribution of the Trojan noise is also required.
On the whole, these challenges make it a non-trivial task to perform Trojan attacks on diffusion models.

\section{Conclusion}
In this paper, we propose the first Trojan attack against diffusion models with diverse targets and triggers.
Extensive experiments on two benchmark datasets against two diffusion models have demonstrated the effectiveness of the proposed attack in terms of six evaluation metrics.

\noindent
\textbf{Acknowledgement.}
This project is partially supported by NSF CCF 1910100, NSF CNS 2046726, C3 AI, NASA ULI, and the Alfred P. Sloan Foundation.






{\small
\bibliographystyle{unsrt} 
\bibliography{egbib}
}

\clearpage
\appendix
\section{More algorithmic details}

\subsection{Details of attacking DDPM}
\label{ap_attack_ddpm}
\subsubsection{Trojan diffusion process}
\noindent
\textbf{How to obtain property of $\bm{k_t}$ (\ie. Equation \ref{eq_k_t}).}
According to $\tilde{q}(x_t|x_{t-1})$ which is defined in Equation \ref{diffusion_transition},
{\small
\begin{flalign}
    &x_t = \sqrt{\alpha_t}x_{t-1} + k_t\mu + \sqrt{1-\alpha_t}\gamma\epsilon_t,\\
    &x_{t-1} =  \sqrt{\alpha_{t-1}}x_{t-2} + k_{t-1}\mu + \sqrt{1-\alpha_{t-1}}\gamma\epsilon_{t-1}. &
\end{flalign}}

\noindent
Hence, $x_t$ could be represented as:
{\small
\begin{align}
    \begin{split}
        &x_t = \sqrt{\alpha_t}(\sqrt{\alpha_{t-1}}x_{t-2}+ k_{t-1}\mu + \sqrt{1-\alpha_{t-1}}\gamma\epsilon_{t-1})\\
        &+ k_t\mu + \sqrt{1-\alpha_t}\gamma\epsilon_t,
    \end{split}\\
    &= \sqrt{\alpha_t\alpha_{t-1}}x_{t-2} + (k_t + \sqrt{\alpha_t}k_{t-1})\mu + \sqrt{1-\alpha_t\alpha_{t-1}}\gamma\bar{\epsilon}_{t-1},
\end{align}}

\noindent
since $\sqrt{\alpha_t(1-\alpha_{t-1})}\epsilon_{t-1} + \sqrt{1-\alpha_t}\epsilon_t$ could be represented by $\sqrt{1-\alpha_t\alpha_{t-1}}\bar{\epsilon}_{t-1}$. Similarly,
{\small
\begin{align}
    \begin{split}
        &x_t = \sqrt{\alpha_t\alpha_{t-1}\alpha_{t-2}}x_{t-3} + (k_t+\sqrt{\alpha_t}k_{t-1}+\sqrt{\alpha_t\alpha_{t-1}}k_{t-2})\mu \\
        &+ \sqrt{1-\alpha_t\alpha_{t-1}\alpha_{t-2}}\gamma\bar{\epsilon}_{t-2}
    \end{split}\\
    \begin{split}
        &=\dots=\sqrt{\bar{\alpha}_t}x_0 + \sqrt{1-\bar{\alpha}_t}\gamma\epsilon\\
        &+ (k_t + \sqrt{\alpha_t}k_{t-1} + \sqrt{\alpha_t\alpha_{t-1}}k_{t-2} + \dots + \sqrt{\alpha_t\dots\alpha_2}k_1)\mu
    \end{split}
\end{align}}

\noindent
Considering the form of $x_t$ which is shown in Equation \ref{x_t_by_x_0_bd}, we could obtain $\sqrt{1-\bar{\alpha}_t}=k_t + \sqrt{\alpha_t}k_{t-1} + \sqrt{\alpha_t\alpha_{t-1}}k_{t-2} + \dots + \sqrt{\alpha_t\dots\alpha_2}k_1$, \ie, Equation \ref{eq_k_t}.

\noindent
\textbf{How to calculate values of $\bm{k_t}$.}
According to Equation \ref{eq_k_t}, $k_t + \sqrt{\alpha_t}k_{t-1} + \sqrt{\alpha_t\alpha_{t-1}}k_{t-2} + \dots + \sqrt{\alpha_t\dots\alpha_2}k_1 = \sqrt{1-\bar{\alpha}_t}$. Thus, 
{\small
\begin{equation*}
    \begin{split}
        &t=1: k_1 = \sqrt{1-\bar{\alpha}_1},\\
        &t=2: k_2 = \sqrt{1-\bar{\alpha}_2} - \sqrt{\alpha_2}k_1,\\
        &t=3: k_3 = \sqrt{1-\bar{\alpha}_3} - \sqrt{\alpha_3}k_2 - \sqrt{\alpha_3\alpha_2}k_1,\\
        &\dots\\
        &t=T: k_T = \sqrt{1-\bar{\alpha}_T} - \sqrt{\alpha_T}k_{T-1} - \dots - \sqrt{\alpha_T\dots\alpha_2}k_1.
    \end{split}
\end{equation*}}

\noindent
Therefore, $k_{t+1}$ could be derived from $k_t$, and we can calculate values of $\bm{k_t}$ from $t=1$ to $t=T$.

\subsubsection{Trojan training}
\noindent
\textbf{How to obtain $\bm{\tilde{\mu}_q(x_t,x_0)}$ and $\bm{\tilde{\beta}_q(x_t,x_0)}$ (\ie. Equation \ref{eq_mu_q}, \ref{eq_beta_q}).}
According to Equation \ref{eq_exp},
{\small
\begin{align}
    &\tilde{q}(x_{t-1}|x_t, x_0)\propto\exp\{a\cdot x_{t-1}^2 + b\cdot x_{t-1} + C(x_t,x_0) \},
\end{align}}

\noindent
where $a = -\frac{1}{2\gamma^2}(\frac{1}{1-\bar{\alpha}_{t-1}}+\frac{\alpha_t}{\beta_t})$, $b = \frac{1}{\gamma^2}[\frac{\sqrt{\bar{\alpha}_{t-1}}x_0+\sqrt{1-\bar{\alpha}_{t-1}}\mu}{1-\bar{\alpha}_{t-1}} + \frac{\sqrt{\alpha_t}(x_t-k_t\mu)}{1-\alpha_t}]$ and $C(x_t,x_0)$ is an item which does not include $x_{t-1}$.
Hence, the mean and variance of $\tilde{q}(x_{t-1}|x_t, x_0)$ are shown as:
{\small
\begin{align}
    \begin{split}
        &\tilde{\mu}_q(x_t,x_0) = -\frac{b}{2a} = \frac{\sqrt{\alpha_t}(1-\bar{\alpha}_{t-1})}{1-\bar{\alpha}_t}x_t + \frac{\sqrt{\bar{\alpha}_{t-1}}\beta_t}{1-\bar{\alpha}_t}x_0\\
        &+ \frac{\sqrt{1-\bar{\alpha}_{t-1}}\beta_t - \sqrt{\alpha_t}(1-\bar{\alpha}_{t-1})k_t}{1-\bar{\alpha}_t}\mu,\\
    \end{split}\\
    &\tilde{\beta}_q(x_t,x_0) = -\frac{1}{2a} = \frac{(1-\bar{\alpha}_{t-1})\beta_t}{1-\bar{\alpha}_t}\gamma^2.
\end{align}}

\section{More implementation details}
\label{ap_imple_detail}
Following \cite{ddpm}, we model $\epsilon_\theta$ using the U-Net \cite{unet} which is based on a Wide ResNet \cite{wideresnet}, where the parameters $\theta$ are shared across time. The pre-trained diffusion models on CIFAR-10 and CelebA datasets are downloaded from \url{https://github.com/pesser/pytorch_diffusion} and \url{https://github.com/ermongroup/ddim}, respectively.
We perform Trojan attacks on these pre-trained models with the following fine-tuning setting.
We set the learning rate as $2\times 10^{-4}$ without any sweeping and use Adam \cite{adam} as the optimizer.
Besides, we adopt the same number of training steps and variance schedule as in \cite{ddpm}, \ie, $T=1000$ and $\{\beta_i\}_{i=1}^T$ are constants increasing linearly from $\beta_1=1\times 10^{-4}$ to $\beta_T=0.02$.
In particular, we set $\eta=0$ and $S=100$ in DDIM since it performs well with this setting based on both sampling speed and sampling quality according to \cite{ddim}.
In addition, we also study the effect of $\eta$ and $S$ on the attack performance of Trojaned DDIMs in Appendix \ref{ap_eta}.
Moreover, as suggested in \cite{ddim}, the strided sampling procedure $\{\tau\}_{i=1}^S$ in DDIM is configured in a quadratic way (\ie. $\tau_i = \lfloor ci\rfloor$ for some $c$) on CIFAR-10 dataset and in a linear way (\ie. $\tau_i = \lfloor ci^2\rfloor$ for some $c$) on CelebA dataset.

In each training step, we load a batch of training data. Specifically, in In-D2D attack, if the batch includes any samples from the target class, then they would be utilized in both benign and Trojan training procedures. Otherwise, the batch is only used in benign training.
By contrast, in Out-D2D attack and D2I attack, since the adversarial targets do not exist in the data distribution, we additionally construct a target loader which consists of data from the target distribution, \ie, all training samples from class 8 in MNIST dataset (Out-D2D attack) and the Mickey Mouse image (D2I attack). Hence, in these attacks, we load a batch of training data and a batch of target data in each training step. The target data are only used in the Trojan training procedure. In particular, the batch size of the target data is 50\% and 10\% smaller than that of the training data in Out-D2D attack and D2I attack, respectively, since reversing the Gaussian distribution to another distribution instead of a specific image is more challenging.

\section{More details of evaluation metrics}
\label{ap_metric}
\subsection{Evaluation metrics for benign performance}
\noindent
\textbf{FID.}
We adopt the Frechet Inception Distance (FID) defined in \cite{fid}, which reflects the quality and the diversity of the generated images.

\noindent
\textbf{Precision and recall.}
We adopt the precision and recall defined in \cite{precision}, which separately reflect the quality and the diversity of the generated images. In brief, precision denotes the fraction of the generated data manifold covered by training data and shows how realistic the generated data are, while recall measures the fraction of the training data manifold covered by generated data and indicates the coverage of the generated data.

\subsection{Evaluation metrics for attack performance}
\noindent
\textbf{Attack precision.}
Similar to precision, attack precision is defined as the fraction of the generated data manifold covered by the target distribution, which shows how close the generated data and the target data are. 
Specifically, in In-D2D attack, the target data are training samples from class 8 (\textit{horse}) on CIFAR-10 dataset while training samples from class 8 (\textit{faces with heavy makeup, with mouth slightly open, with smiling}) on CelebA dataset.
And in Out-D2D attack, the target data are training samples from class 8 (\textit{handwritten eight}) on MNIST dataset.

\noindent
\textbf{ASR.}
Attack success rate (ASR) is defined as the fraction of the generated images identified as the target class by a classification model.
Specifically, in In-D2D attack, we train a ResNet18 \cite{resnet18} of 93.36\% testing accuracy on CIFAR-10 dataset. Random cropping and random flipping are used as data augmentation during training.
Besides, we train  a ResNet18 \cite{resnet18} of 80.24\% testing accuracy on CelebA. Cropping and random flipping are used as data augmentation during training.
In Out-D2D attack on both datasets, we train a simple network proposed in \cite{wanet} with 99.56\% testing accuracy on MNIST dataset. Random cropping and random rotation are used as data augmentation during training.

\noindent
\textbf{MSE.}
Mean square error (MSE) is measured between the generated images and the target image, \ie. Mickey Mouse, which indicates how similar these images are. A smaller MSE corresponds to a higher similarity between them.

\noindent
\textbf{Remark.}
Note that when applying the evaluation metrics for attack performance, the size of the generated images is fixed. Instead, the size of the images used for comparison 
(\ie. the target data)
is scaled to the same size as the generated images (\ie, 32$\times$32 on CIFAR-10 dataset and 64$\times$64 on CelebA dataset).

\section{More ablation studies}
\subsection{Effect of patch size in patch-based attack}
\label{ap_ablation_ps}
In this part, we aim to explore how the size of the patch trigger influences the attack performance of Trojaned diffusion models under patch-based attacks.

\begin{figure}[htbp]
    \centering
    \includegraphics[width = 0.35\textwidth]{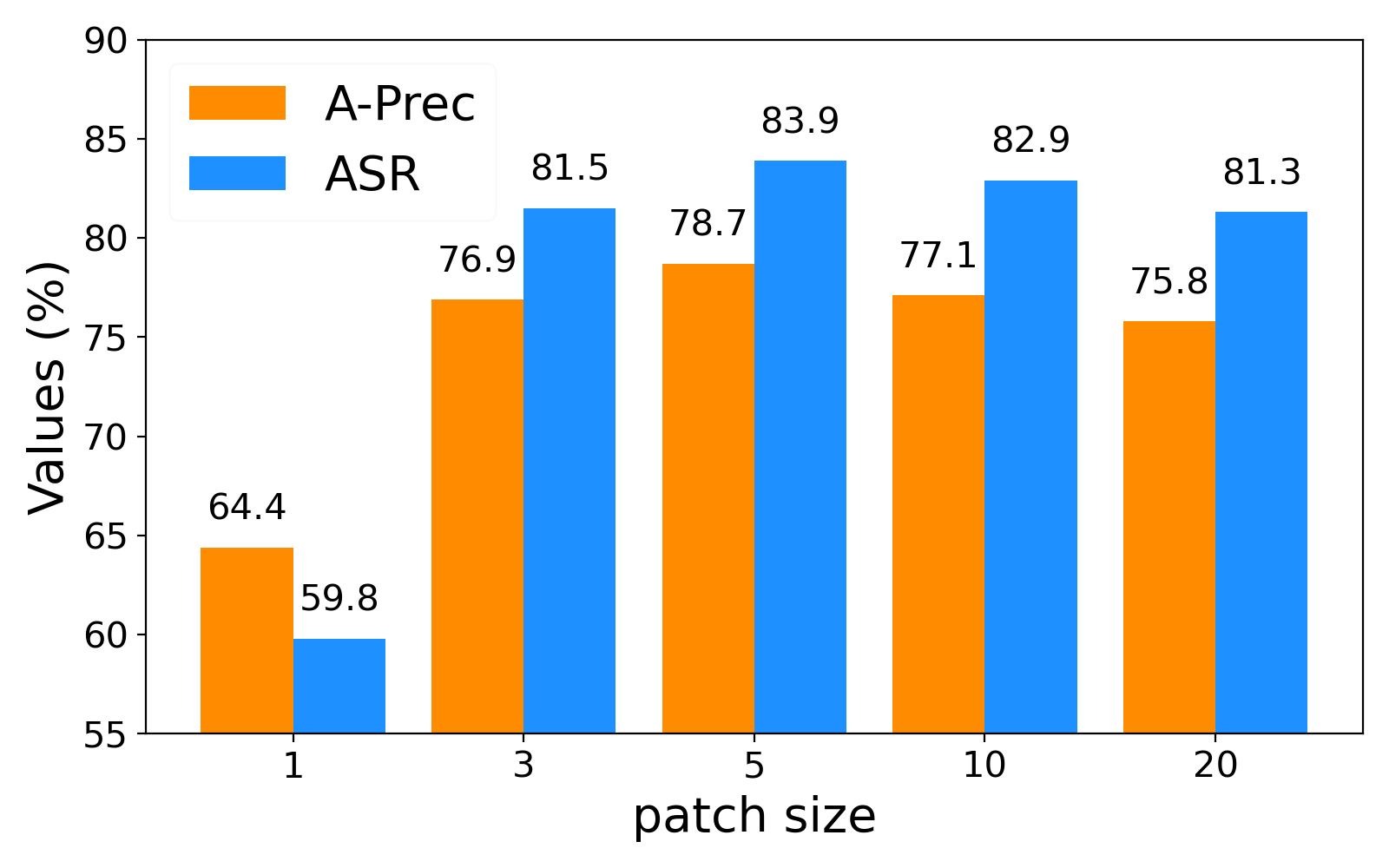}
    \caption{Attack performance against DDIMs under patch-based In-D2D attack on CIFAR-10 dataset with different sizes of the patch.}
    \label{fig_effect_ps}
\end{figure}

As shown in Figure \ref{fig_effect_ps}, a moderate patch size is desired in terms of the two metrics.
Similar to the analysis in Section 4.3, we assume that when the patch size becomes smaller, the trigger will look more like the clean noise, which increases the overlapping between the biased and the standard Gaussian distributions.
If the patch size is smaller to a certain extent (\eg, patch size = 1), it is hard for the model to identify between clean noise and Trojan noise during training,
thus learning a bad Trojaned diffusion model.
Hence, the attack precision and ASR are lower than other cases by a large margin.

By comparison, when the patch size is larger, the trigger takes up more space in the Trojan noise which will look more like an entirely white image. Since we adopt $\gamma=0.1$ on the patch as mentioned at the end of Section 3, \ie, there is still a small extent of noise on the patch, the Trojan noise is still capable of providing sufficient random space for learning a Trojaned diffusion model even with a large patch size. Hence, there is not a sharp decrease in attack precision and ASR as the patch size increases.
In conclusion, except for the extremely small size, the proposed \name is still robust to different sizes of patch under patch-based attacks.




\subsection{Effect of $\eta$ and $S$ in Trojaned DDIMs}
\label{ap_eta}
As mentioned in Appendix \ref{ap_imple_detail}, we set $\eta=0$ and $S=100$ in DDIM since it performs well with this setting considering both the sampling speed and the quality of the generated images according to \cite{ddim}, which has discussed the effect of $\eta$ and $S$ on the benign performance on DDIMs.
In this part, we focus on how the settings of $\eta$ and $S$ affect the attack performance against DDIMs. 

\noindent
\textbf{Effect of $\bm{\eta}$.}
Firstly, we explore the effect of $\eta$ on the attack performance against DDIMs. To this end, we fix $S=100$ and vary $\eta$ from 0.0 to 1.0.
As shown in the first row of Table \ref{tab_eta}, the Trojaned DDIMs exhibit consistently high attack performance under different settings of $\eta$. For instance, the ASRs are 87.30\% on average and the variance is down to 1.24\%, which demonstrates that the proposed \name is robust to different settings of $\eta$ when attacking DDIMs.

\noindent
\textbf{Effect of S.}
Then, we study the effect of $S$ on the attack performance against DDIMs. Thus, we fix $\eta = 0.0$ and vary $S$ from 10 to 1000. The results are illustrated in the second row of  Table \ref{tab_eta}.
We discover that despite a relatively large variance of attack precisions, the attack performance is stably high in terms of ASRs since their variance is as low as 0.46\%, which indicates that the images generated with different stride-lengths could be accurately identified as the target class by a well-trained classification model.

\begin{table}[htbp]
  \centering
  \scalebox{0.8}{
    \begin{tabular}{l|rrrrr|rr}
\cmidrule{1-7}    $\eta$ & \textbf{0.0 } & \textbf{0.2 } & \textbf{0.5 } & \multicolumn{1}{r|}{\textbf{1.0 }} & \multicolumn{1}{l}{\textbf{Avg}} & \multicolumn{1}{l}{\textbf{Var}} &  \\
\cmidrule{1-7}    \textbf{A-Prec} & 80.00  & 78.70  & 81.90  & \multicolumn{1}{r|}{78.90 } & \multicolumn{1}{r}{79.88 } & 2.15  &  \\
    \textbf{ASR} & 87.00  & 87.90  & 89.50  & \multicolumn{1}{r|}{87.30 } & \multicolumn{1}{r}{87.93 } & 1.24  &  \\
    \midrule
    \textbf{S} & \textbf{10 } & \textbf{20 } & \textbf{50 } & \textbf{100 } & \textbf{1000 } & \multicolumn{1}{l}{\textbf{Avg}} & \multicolumn{1}{l}{\textbf{Var}} \\
    \midrule
    \textbf{A-Prec} & 85.40  & 83.70  & 78.90  & 78.90  & 77.90  & 80.96  & 11.27  \\
    \textbf{ASR} & 86.30  & 86.20  & 85.40  & 87.30  & 86.40  & 86.32  & 0.46  \\
    \bottomrule
    \end{tabular}}
  \caption{Attack performance (\%) against DDIMs under blend-based In-D2D attack on CIFAR-10 dataset with different $\eta$ and $S$.}
  \label{tab_eta}%
\end{table}%

\section{More experimental results}
\label{ap_result}
In this section, we aim to answer: \textit{Why does the fine-tuned DDPM suffer a rise in FID on CIFAR-10 dataset as shown in Table \ref{tab_ddpm}, compared to the pre-trained model?}

\begin{figure}[htbp]
    \centering
    \includegraphics[width = 0.35\textwidth]{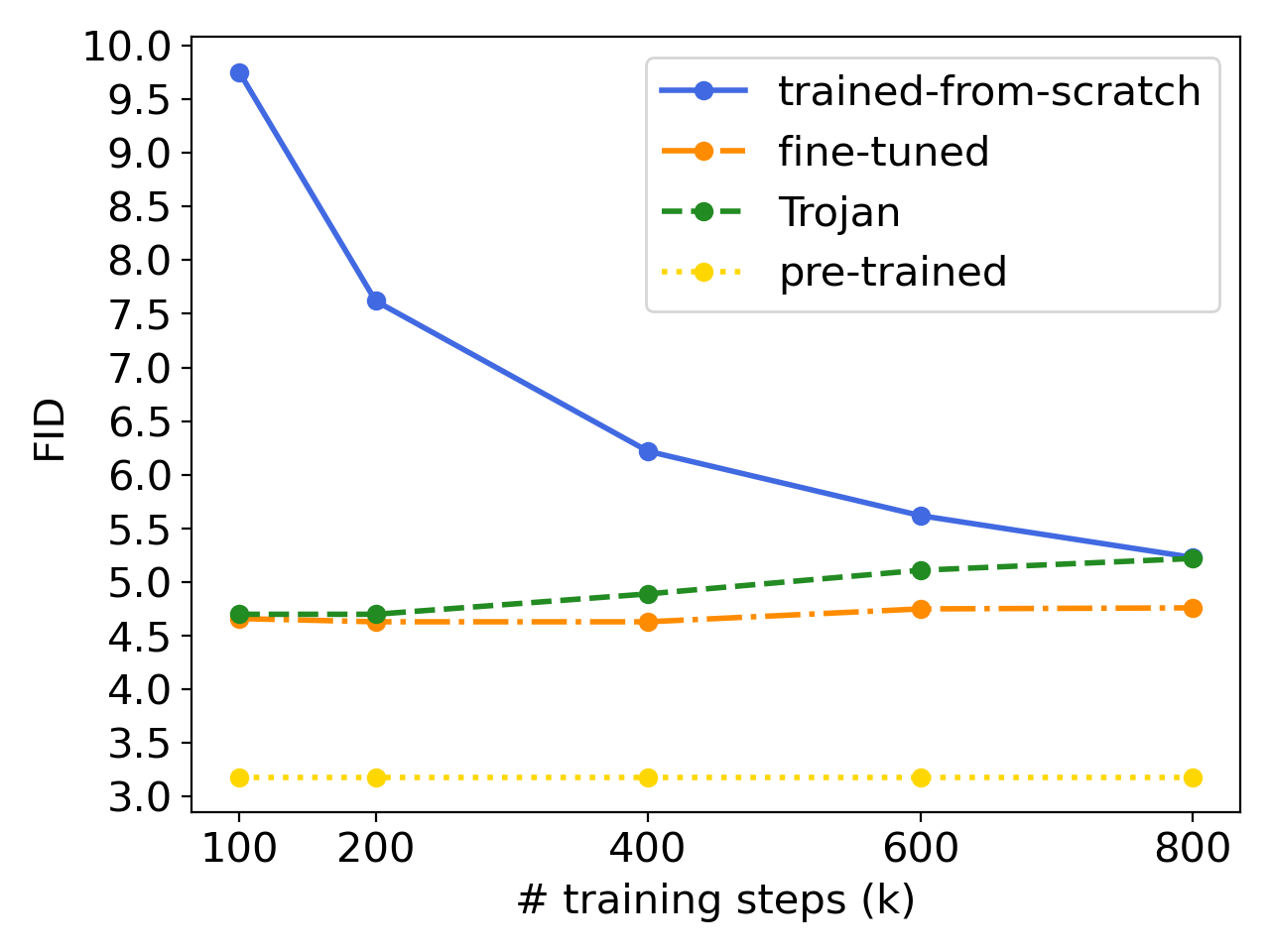}
    \caption{FID of different DDPMs over the training steps on CIFAR-10 dataset.}
    \label{ddpm_fid_drop}
\end{figure}

According to \cite{ddpm}, it requires 800k steps to train a DDPM on CIFAR-10 dataset. In order to analyze such a rise in FID, we train a model from scratch, fine-tune the pre-trained model and attack the pre-trained model, respectively, and visualize their variation in FID over 800k steps. 
We analyze the results, which are shown in Figure \ref{ddpm_fid_drop}, from three perspectives.

Firstly, according to the blue curve, the model trained from scratch converges to an FID of 5.23. It demonstrates that based on the open-source PyTorch code mentioned in Appendix \ref{ap_imple_detail}, the trained model does not achieve the low FID of the pre-trained one, which has been confirmed with the authors. 

Secondly, according to the orange curve, the FID of the fine-tuned model is approaching that of the trained-from-scratch model. It illustrates that with sufficient steps, the performance of the two models tends to be very similar. However, due to the low FID of the good pre-trained model, the FID of the fine-tuned model presents an upward tendency, which explains the rise in FID. 

Finally, according to the green curve, the FIDs of the attacked model and the fine-tuned model are consistently similar, which again confirms our analysis that TrojDiff does not hurt the benign performance.

\section{More visualization results}
\label{ap_visual}

\subsection{Visualization of generated adversarial targets}
\begin{figure*}[htbp]
    \centering
    \begin{subfigure}{0.35\linewidth}
        \includegraphics[width = 1.0\textwidth]{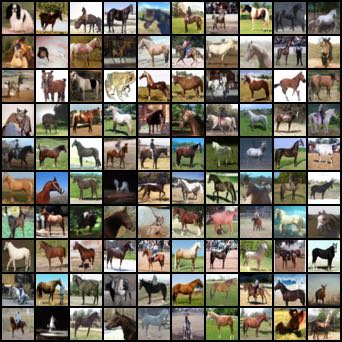}
        \caption{In-D2D attack (CIFAR-10)}
    \end{subfigure}
    \begin{subfigure}{0.35\linewidth}
        \includegraphics[width = 1.0\textwidth]{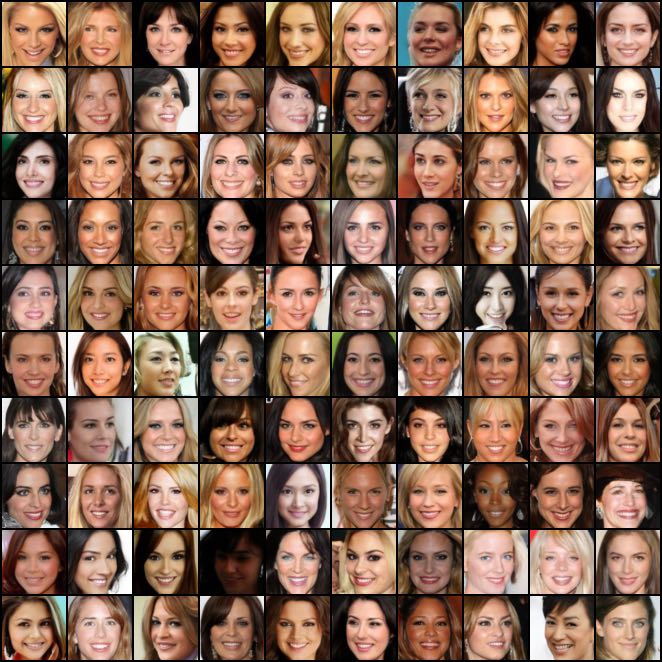}
        \caption{In-D2D attack (CelebA)}
    \end{subfigure}
     \begin{subfigure}{0.35\linewidth}
        \includegraphics[width = 1.0\textwidth]{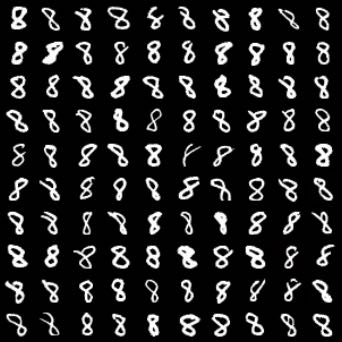}
        \caption{Out-D2D attack (CIFAR-10)}
    \end{subfigure}
    \begin{subfigure}{0.35\linewidth}
        \includegraphics[width = 1.0\textwidth]{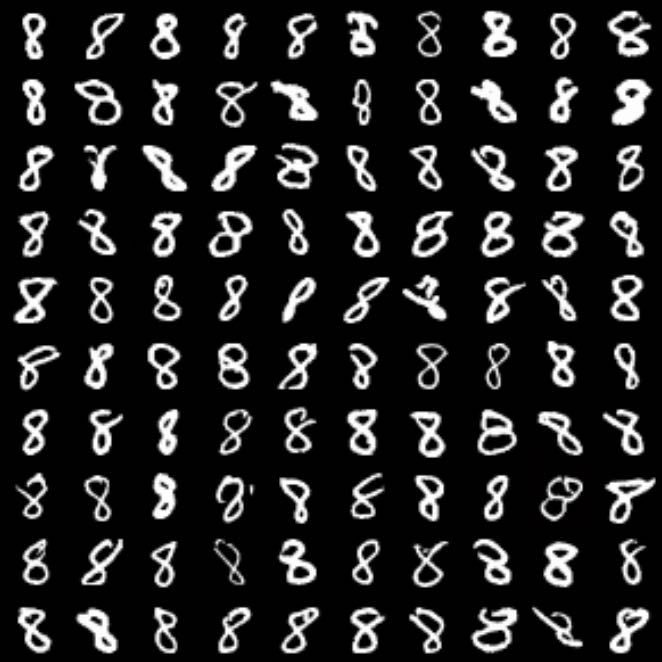}
        \caption{Out-D2D attack (CelebA)}
    \end{subfigure}
     \begin{subfigure}{0.35\linewidth}
        \includegraphics[width = 1.0\textwidth]{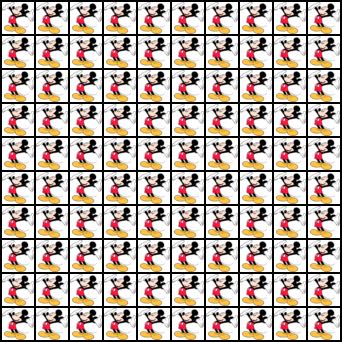}
        \caption{D2I attack (CIFAR-10)}
    \end{subfigure}
    \begin{subfigure}{0.35\linewidth}
        \includegraphics[width = 1.0\textwidth]{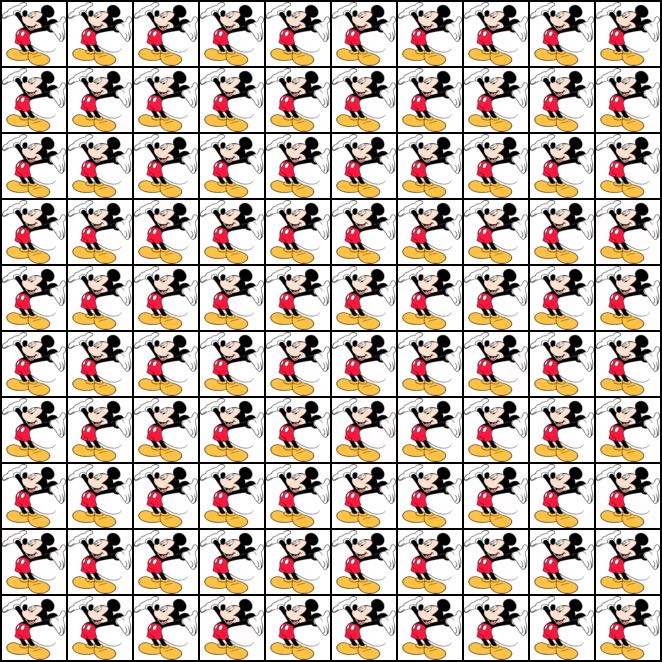}
        \caption{D2I attack (CelebA)}
    \end{subfigure}
    \caption{Adversarial targets generated by Trojaned DDPMs using the blend-based trigger on CIFAR-10 and CelebA datasets.}
    \label{fig_ad_1}
\end{figure*}

\begin{figure*}[htbp]
    \centering
    \begin{subfigure}{0.35\linewidth}
        \includegraphics[width = 1.0\textwidth]{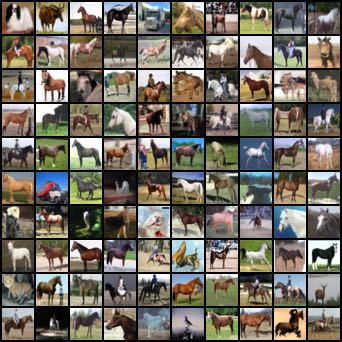}
        \caption{In-D2D attack (CIFAR-10)}
    \end{subfigure}
    \begin{subfigure}{0.35\linewidth}
        \includegraphics[width = 1.0\textwidth]{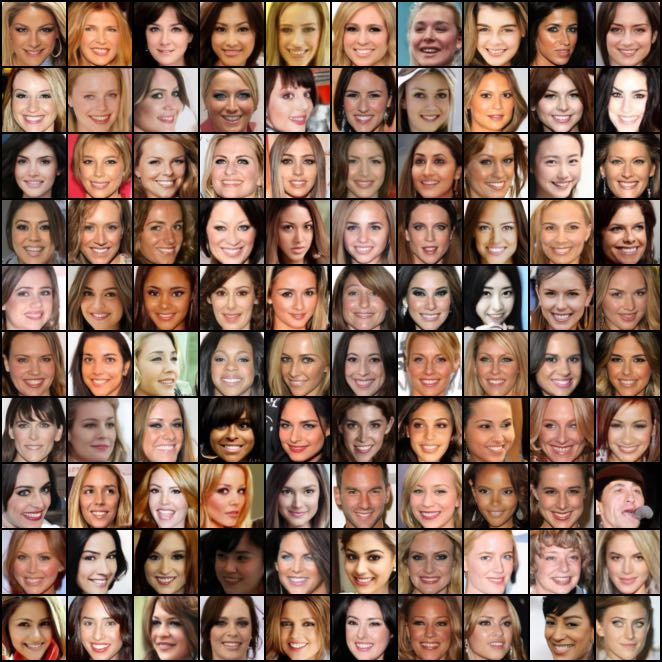}
        \caption{In-D2D attack (CelebA)}
    \end{subfigure}
    \begin{subfigure}{0.35\linewidth}
        \includegraphics[width = 1.0\textwidth]{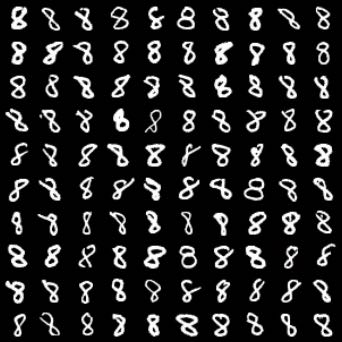}
        \caption{Out-D2D attack (CIFAR-10)}
    \end{subfigure}
    \begin{subfigure}{0.35\linewidth}
        \includegraphics[width = 1.0\textwidth]{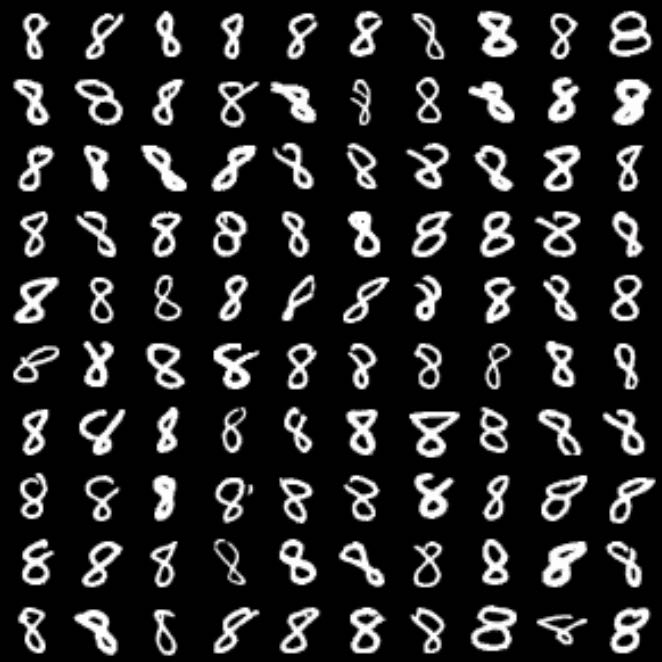}
        \caption{Out-D2D attack (CelebA)}
    \end{subfigure}
    \begin{subfigure}{0.35\linewidth}
        \includegraphics[width = 1.0\textwidth]{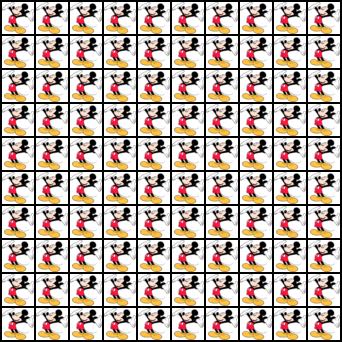}
        \caption{D2I attack (CIFAR-10)}
    \end{subfigure}
    \begin{subfigure}{0.35\linewidth}
        \includegraphics[width = 1.0\textwidth]{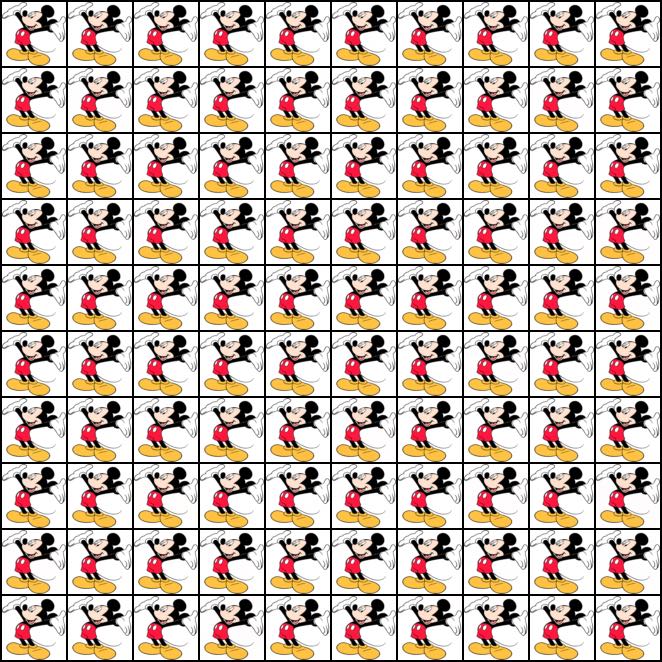}
        \caption{D2I attack (CelebA)}
    \end{subfigure}
    \caption{Adversarial targets generated by Trojaned DDPMs using the patch-based trigger on CIFAR-10 and CelebA datasets.}
    \label{fig_ad_2}
\end{figure*}

\begin{figure*}[htbp]
    \centering
    \begin{subfigure}{0.35\linewidth}
        \includegraphics[width = 1.0\textwidth]{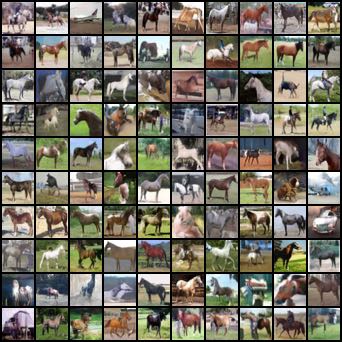}
        \caption{In-D2D attack (CIFAR-10)}
    \end{subfigure}
    \begin{subfigure}{0.35\linewidth}
        \includegraphics[width = 1.0\textwidth]{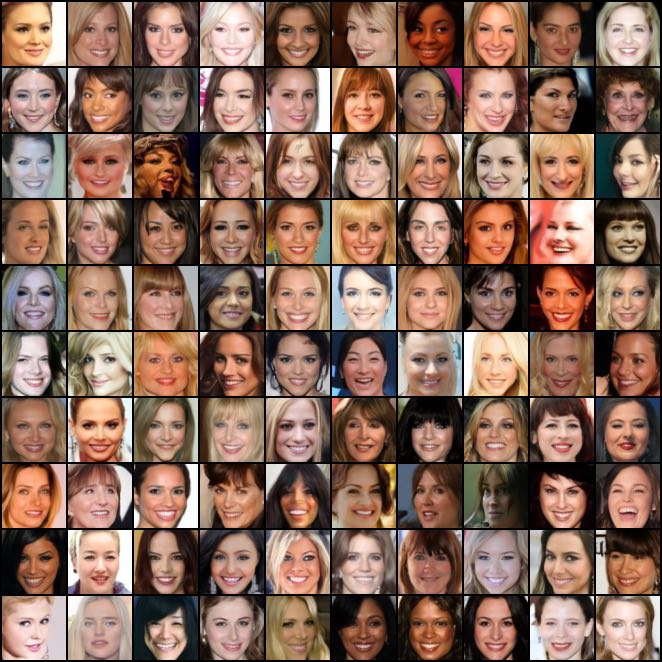}
        \caption{In-D2D attack (CelebA)}
    \end{subfigure}
    \begin{subfigure}{0.35\linewidth}
        \includegraphics[width = 1.0\textwidth]{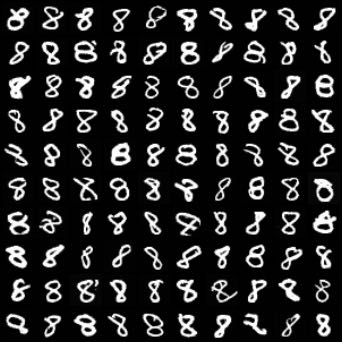}
        \caption{Out-D2D attack (CIFAR-10)}
    \end{subfigure}
    \begin{subfigure}{0.35\linewidth}
        \includegraphics[width = 1.0\textwidth]{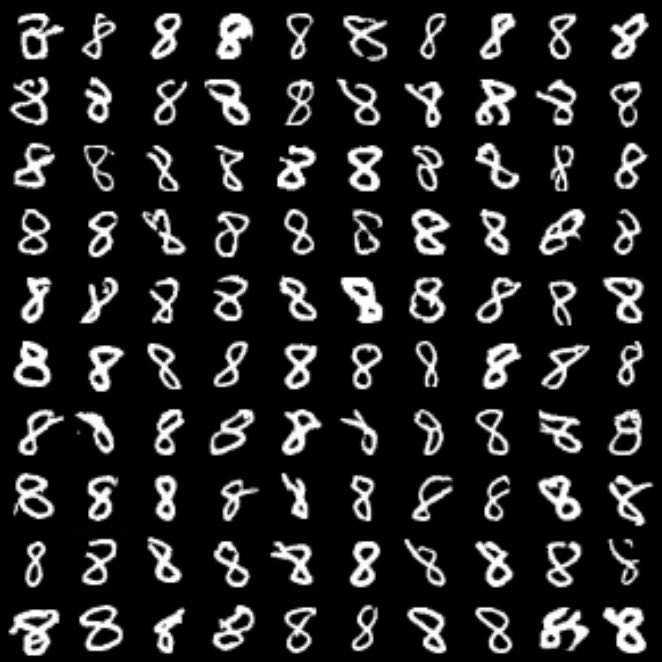}
        \caption{Out-D2D attack (CelebA)}
    \end{subfigure}
    \begin{subfigure}{0.35\linewidth}
        \includegraphics[width = 1.0\textwidth]{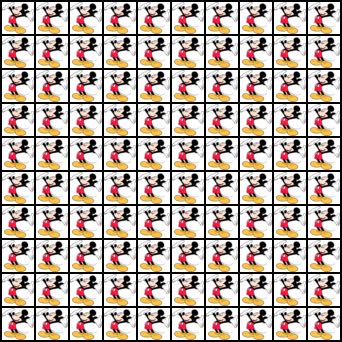}
        \caption{D2I attack (CIFAR-10)}
    \end{subfigure}
    \begin{subfigure}{0.35\linewidth}
        \includegraphics[width = 1.0\textwidth]{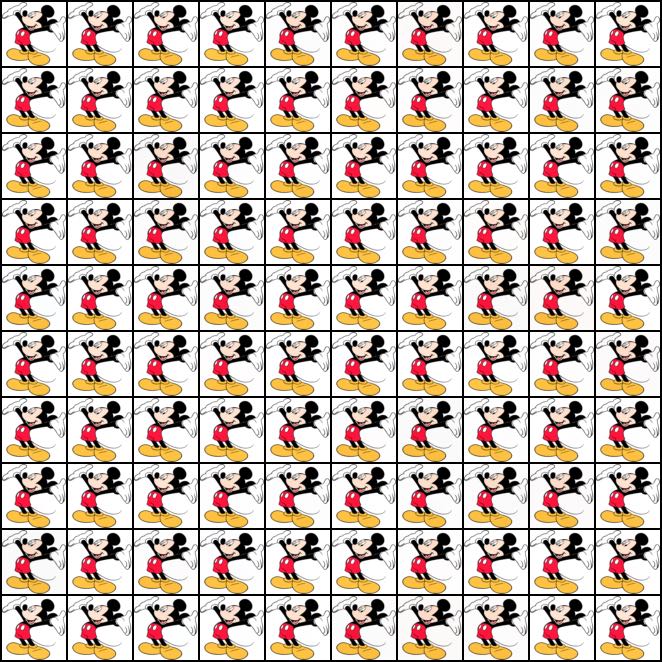}
        \caption{D2I attack (CelebA)}
    \end{subfigure}
    \caption{Adversarial targets generated by Trojaned DDIMs using the blend-based trigger on CIFAR-10 and CelebA datasets.}
    \label{fig_ad_3}
\end{figure*}

\begin{figure*}[htbp]
    \centering
    \begin{subfigure}{0.35\linewidth}
        \includegraphics[width = 1.0\textwidth]{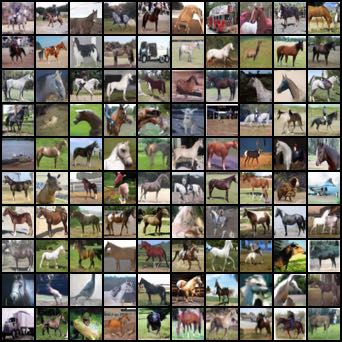}
        \caption{In-D2D attack (CIFAR-10)}
    \end{subfigure}
    \begin{subfigure}{0.35\linewidth}
        \includegraphics[width = 1.0\textwidth]{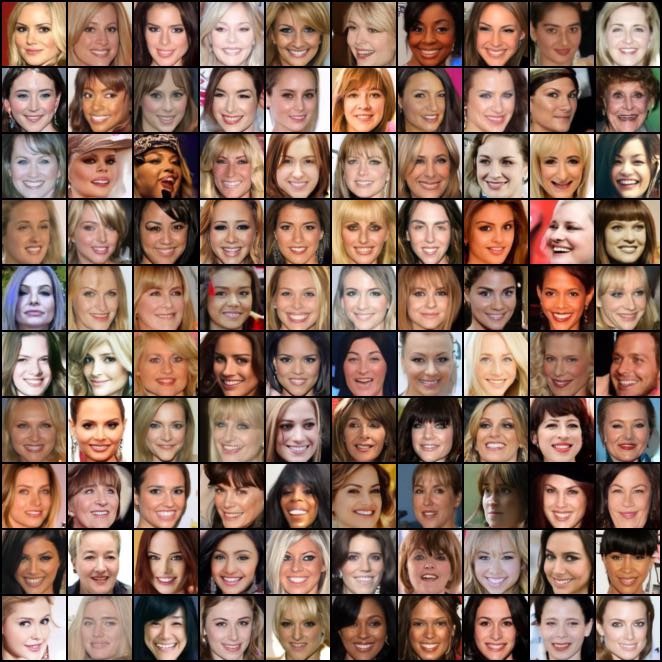}
        \caption{In-D2D attack (CelebA)}
    \end{subfigure}
    \begin{subfigure}{0.35\linewidth}
        \includegraphics[width = 1.0\textwidth]{figures/visual_compress/ddpm_patch_d2dout_cifar10_.jpg}
        \caption{Out-D2D attack (CIFAR-10)}
    \end{subfigure}
    \begin{subfigure}{0.35\linewidth}
        \includegraphics[width = 1.0\textwidth]{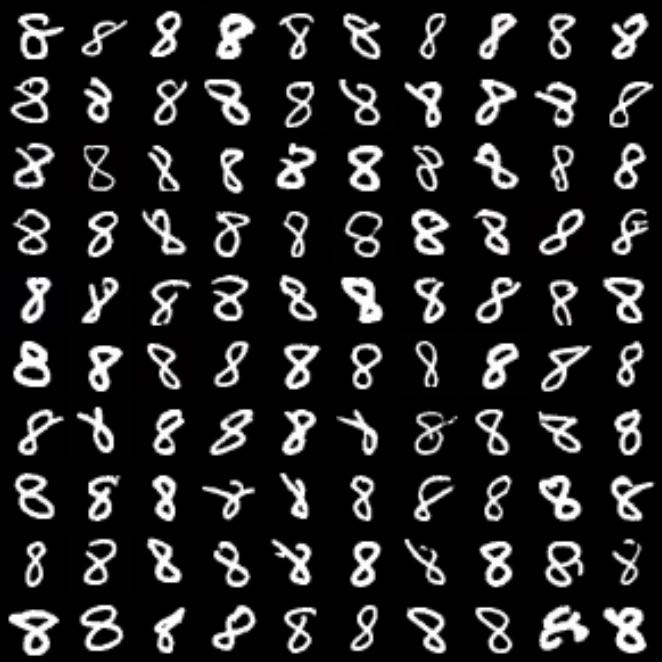}
        \caption{Out-D2D attack (CelebA)}
    \end{subfigure}
    \begin{subfigure}{0.35\linewidth}
        \includegraphics[width = 1.0\textwidth]{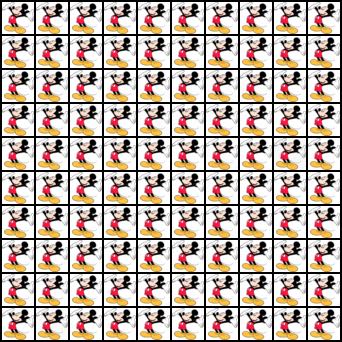}
        \caption{D2I attack (CIFAR-10)}
    \end{subfigure}
    \begin{subfigure}{0.35\linewidth}
        \includegraphics[width = 1.0\textwidth]{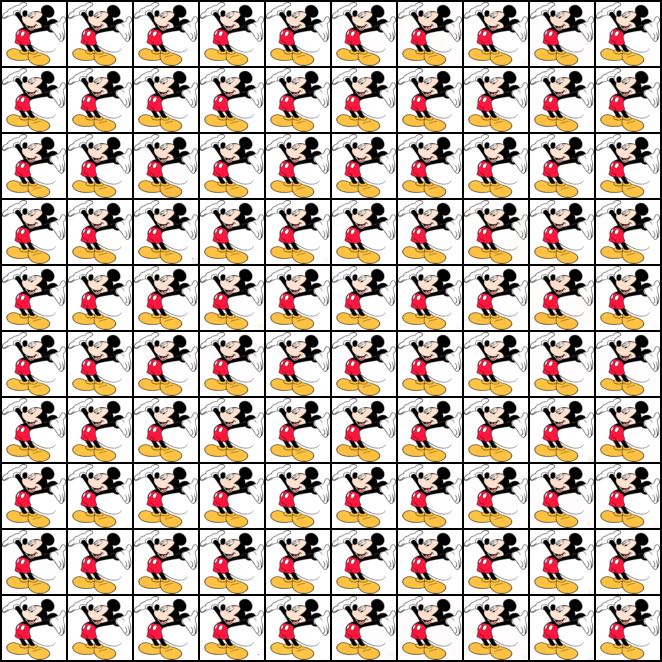}
        \caption{D2I attack (CelebA)}
    \end{subfigure}
    \caption{Adversarial targets generated by Trojaned DDIMs using the patch-based trigger on CIFAR-10 and CelebA datasets.}
    \label{fig_ad_4}
\end{figure*}

Figure \ref{fig_ad_1}-\ref{fig_ad_4} show more adversarial targets randomly generated by Trojaned DDPMs and Trojaned DDIMs under three types of attacks using the blend-based trigger on CIFAR-10 and CelebA datasets. Under In-D2D attacks, the generated adversarial targets could be well aligned to the corresponding target classes, \ie, horse and faces with heavy makeup, mouth slightly open and smiling. Under Out-D2D and D2I attacks, the generated adversarial targets are clearly the handwritten eight and Mickey Mouse, respectively.

\subsection{Visualization of Trojan generative process}
\begin{figure*}[t]
  \centering
  \begin{subfigure}{0.9\linewidth}
    \includegraphics[width = 1.0\textwidth]{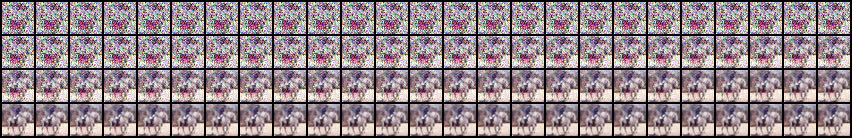}
    \caption{Trojan generative process under In-D2D attack with blend-based trigger.}
  \end{subfigure}
  
  \begin{subfigure}{0.9\linewidth}
    \includegraphics[width = 1.0\textwidth]{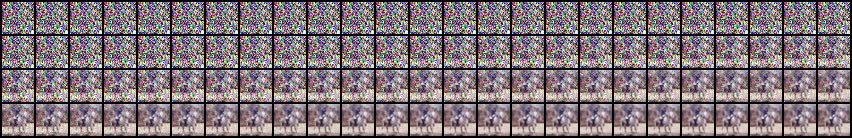}
    \caption{Trojan generative process under In-D2D attack with patch-based trigger.}
  \end{subfigure}
  
  \begin{subfigure}{0.9\linewidth}
    \includegraphics[width = 1.0\textwidth]{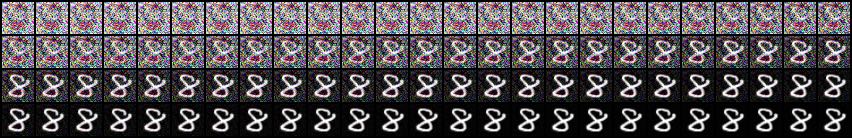}
    \caption{Trojan generative process under Out-D2D attack with blend-based trigger.}
  \end{subfigure}
  
  \begin{subfigure}{0.9\linewidth}
    \includegraphics[width = 1.0\textwidth]{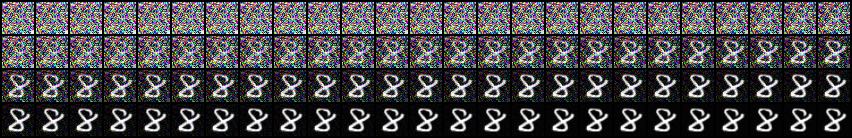}
    \caption{Trojan generative process under Out-D2D attack with patch-based trigger.}
  \end{subfigure}
  
  \begin{subfigure}{0.9\linewidth}
    \includegraphics[width = 1.0\textwidth]{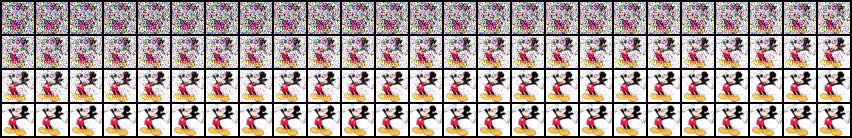}
    \caption{Trojan generative process under D2I attack with blend-based trigger.}
  \end{subfigure}
  
  \begin{subfigure}{0.9\linewidth}
    \includegraphics[width = 1.0\textwidth]{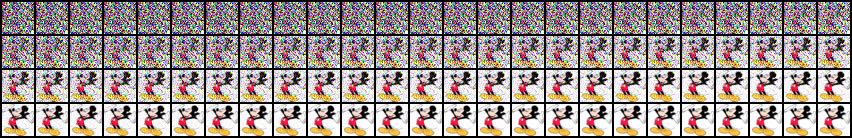}
    \caption{Trojan generative process under D2I attack with patch-based trigger.}
  \end{subfigure}
  \caption{Trojan generative processes of the Trojaned DDIMs under In-D2D, Out-D2D and D2I attacks using two types of triggers on CIFAR-10 dataset.}
  \label{fig_visual_cifar}
\end{figure*}

\begin{figure*}[t]
  \centering
  \begin{subfigure}{0.9\linewidth}
    \includegraphics[width = 1.0\textwidth]{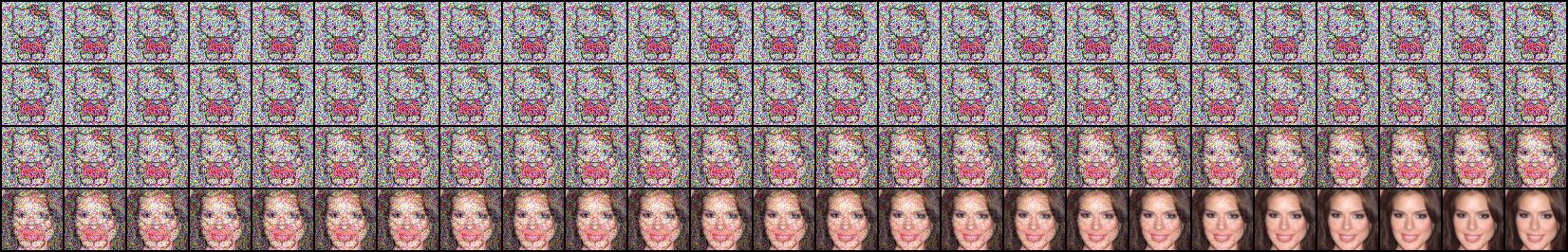}
    \caption{Trojan generative process under In-D2D attack with blend-based trigger.}
  \end{subfigure}
  
  \begin{subfigure}{0.9\linewidth}
    \includegraphics[width = 1.0\textwidth]{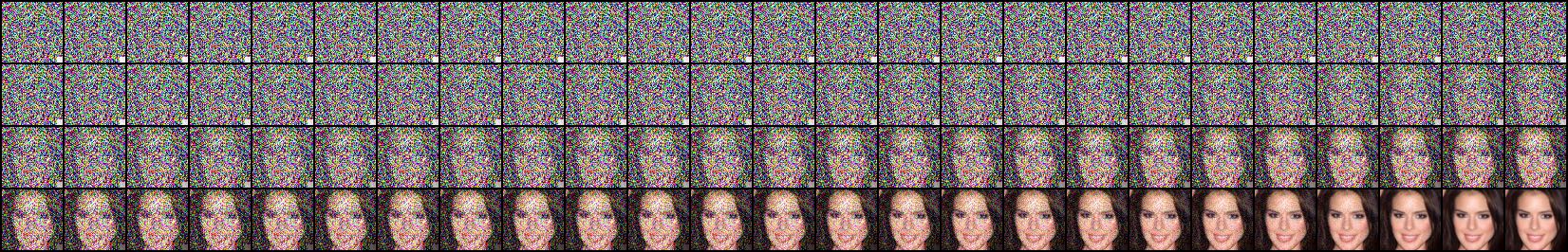}
    \caption{Trojan generative process under In-D2D attack with patch-based trigger.}
  \end{subfigure}
  
  \begin{subfigure}{0.9\linewidth}
    \includegraphics[width = 1.0\textwidth]{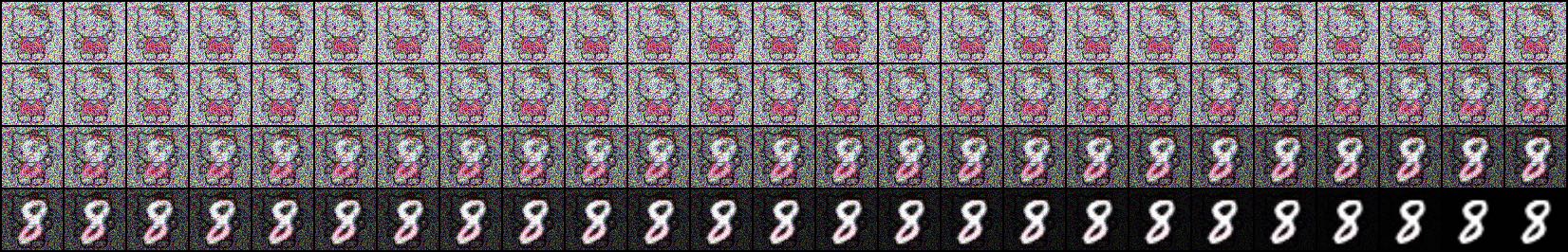}
    \caption{Trojan generative process under Out-D2D attack with blend-based trigger.}
  \end{subfigure}
  
  \begin{subfigure}{0.9\linewidth}
    \includegraphics[width = 1.0\textwidth]{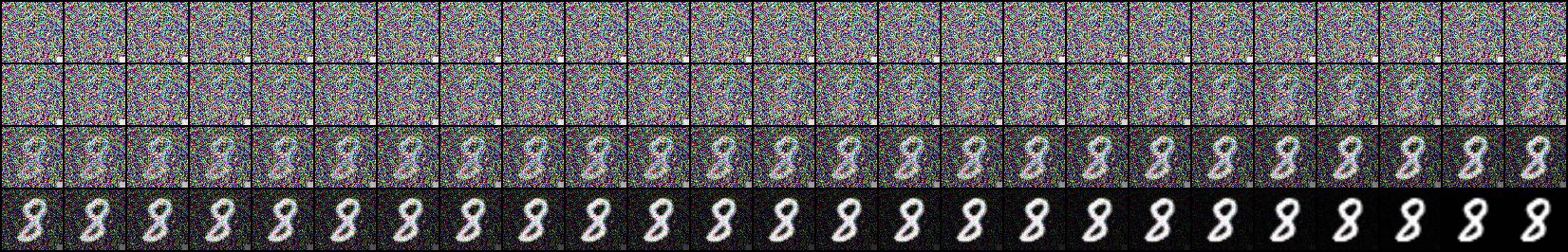}
    \caption{Trojan generative process under Out-D2D attack with patch-based trigger.}
  \end{subfigure}
  
  \begin{subfigure}{0.9\linewidth}
    \includegraphics[width = 1.0\textwidth]{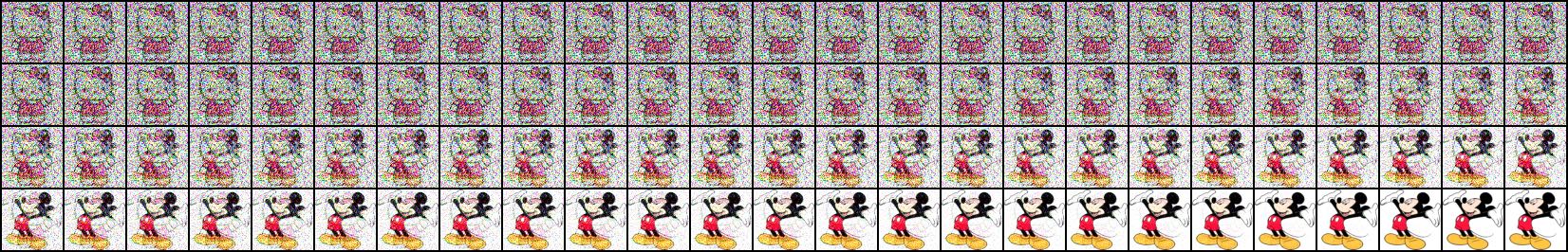}
    \caption{Trojan generative process under D2I attack with blend-based trigger.}
  \end{subfigure}
  
  \begin{subfigure}{0.9\linewidth}
    \includegraphics[width = 1.0\textwidth]{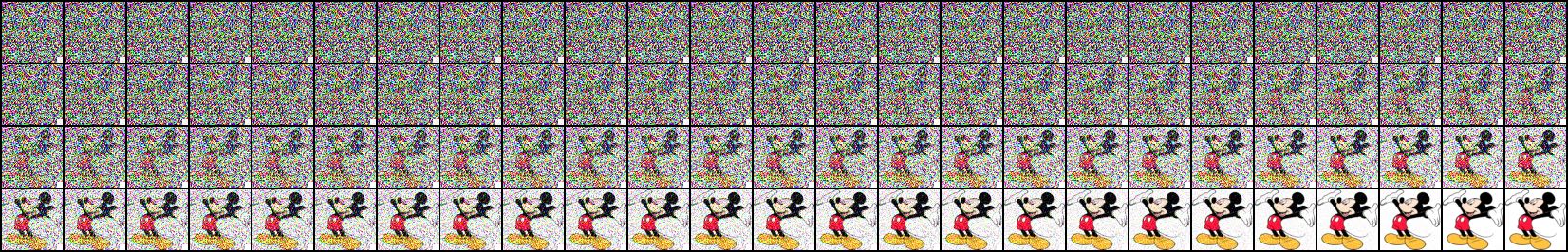}
    \caption{Trojan generative process under D2I attack with patch-based trigger.}
  \end{subfigure}
  \caption{Trojan generative processes of the Trojaned DDIMs under In-D2D, Out-D2D and D2I attacks using two types of triggers on CelebA dataset.}
  \label{fig_visual_celeba}
\end{figure*}

\begin{figure*}[htbp]
  \centering
  \begin{subfigure}{0.9\linewidth}
    \includegraphics[width = 1.0\textwidth]{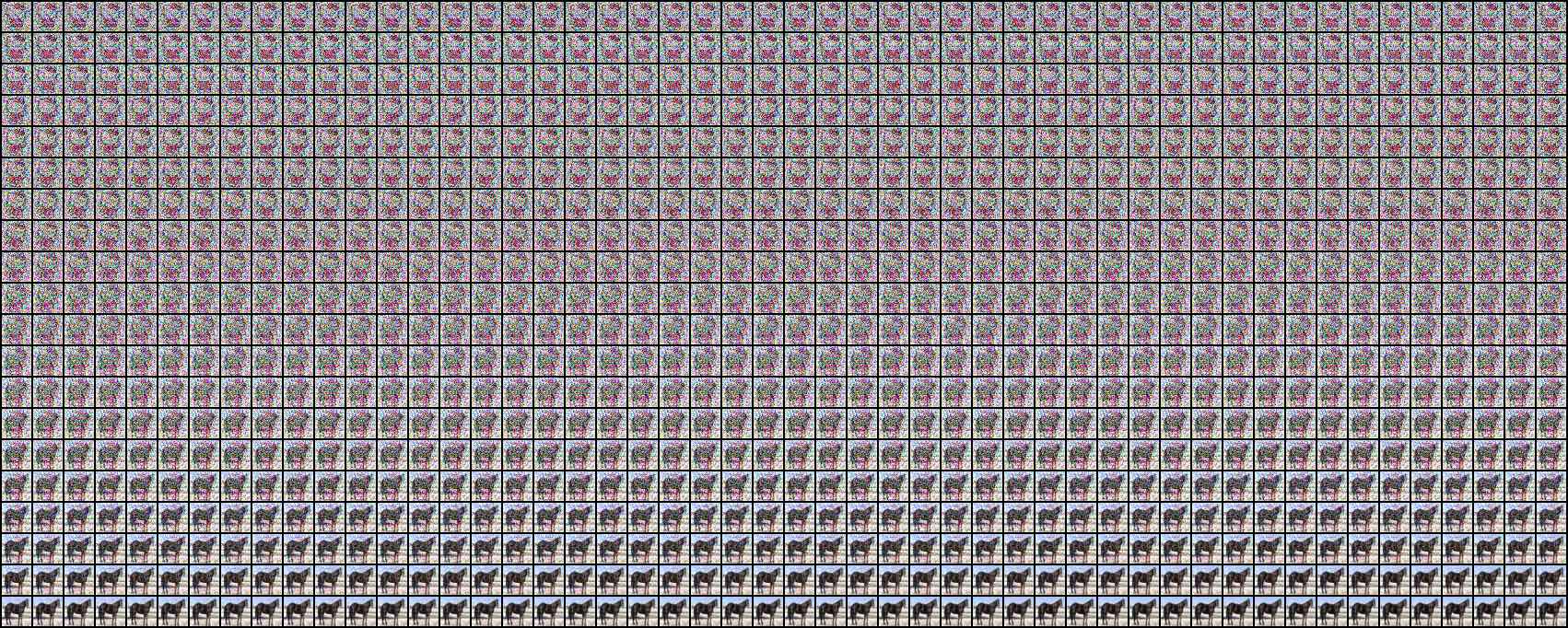}
    \caption{Trojan generative process under In-D2D attack with blend-based trigger.}
  \end{subfigure}
  
  \begin{subfigure}{0.9\linewidth}
    \includegraphics[width = 1.0\textwidth]{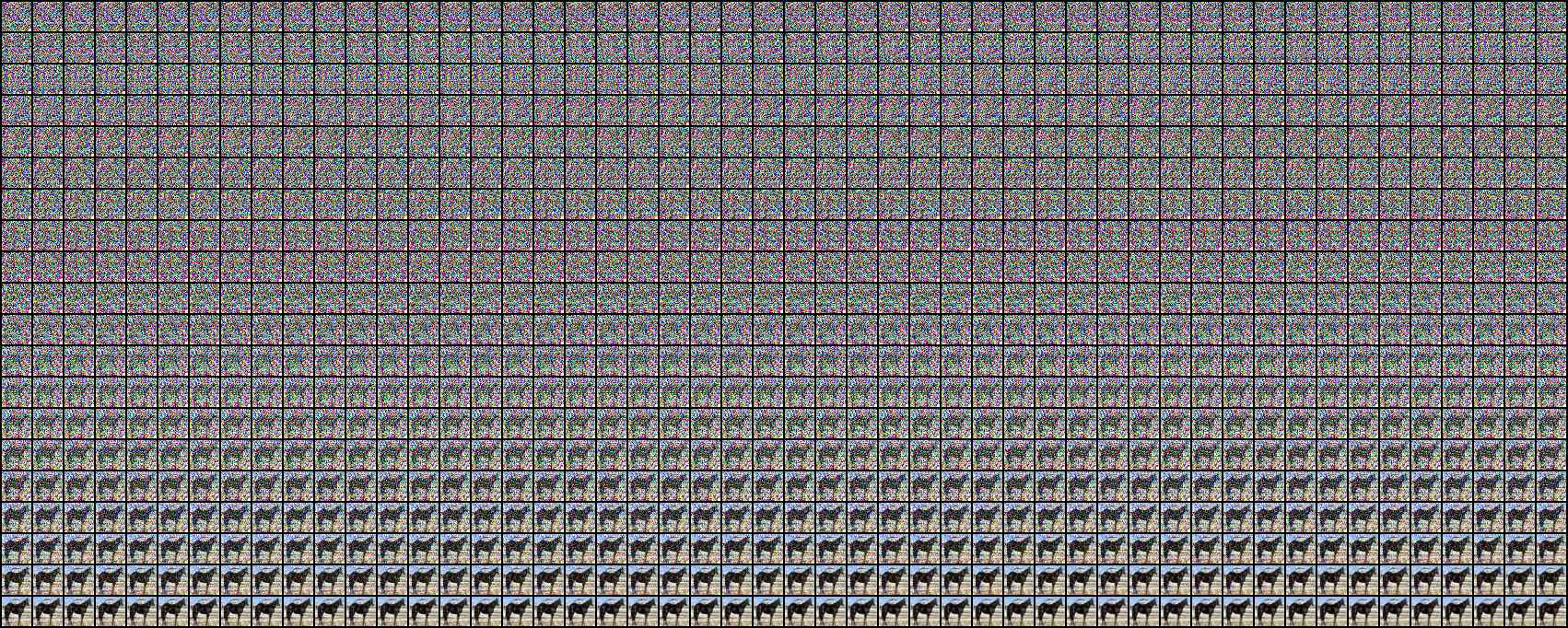}
    \caption{Trojan generative process under In-D2D attack with patch-based trigger.}
  \end{subfigure}
  
  \caption{Trojan generative processes of the Trojaned DDPMs under In-D2D attack using two types of triggers on CIFAR-10 dataset.}
  \label{fig_visual_cifar_ddpm_d2din}
\end{figure*}

\begin{figure*}[htbp]
  \centering
  
  \begin{subfigure}{0.9\linewidth}
    \includegraphics[width = 1.0\textwidth]{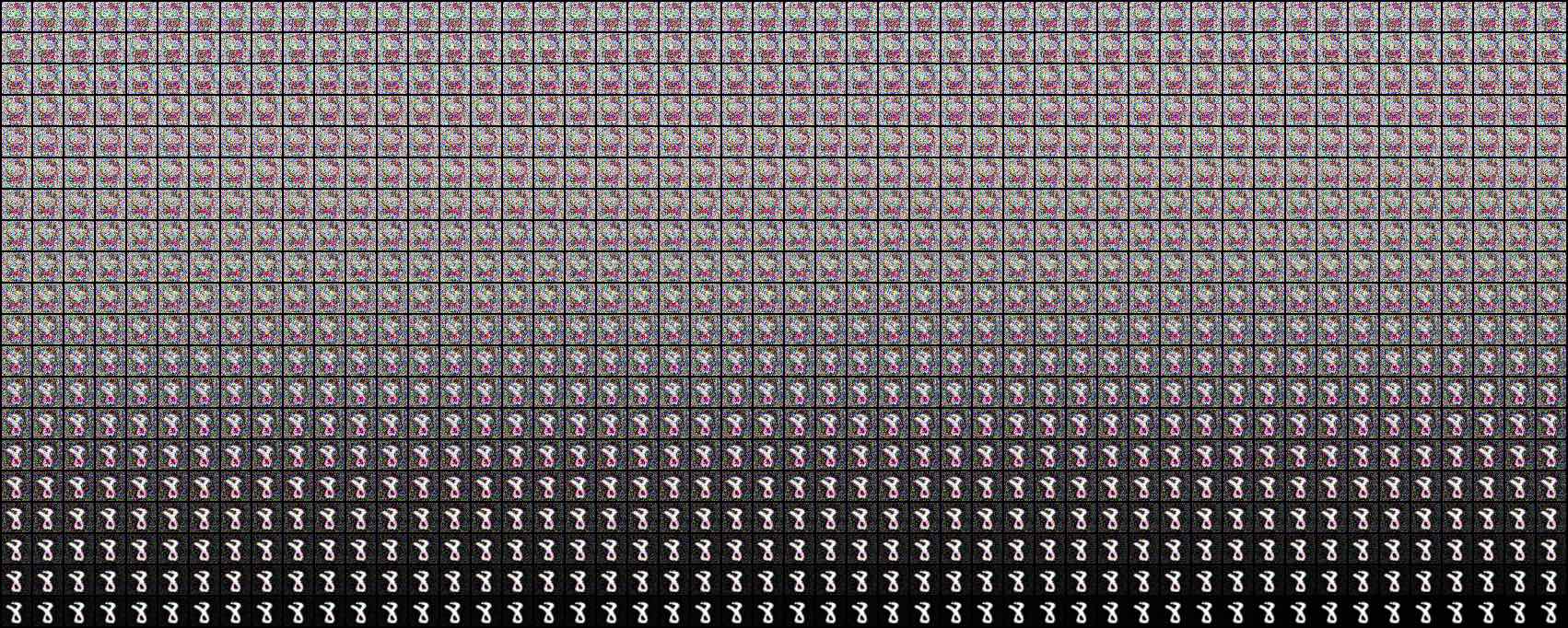}
    \caption{Trojan generative process under Out-D2D attack with blend-based trigger.}
  \end{subfigure}
  
  \begin{subfigure}{0.9\linewidth}
    \includegraphics[width = 1.0\textwidth]{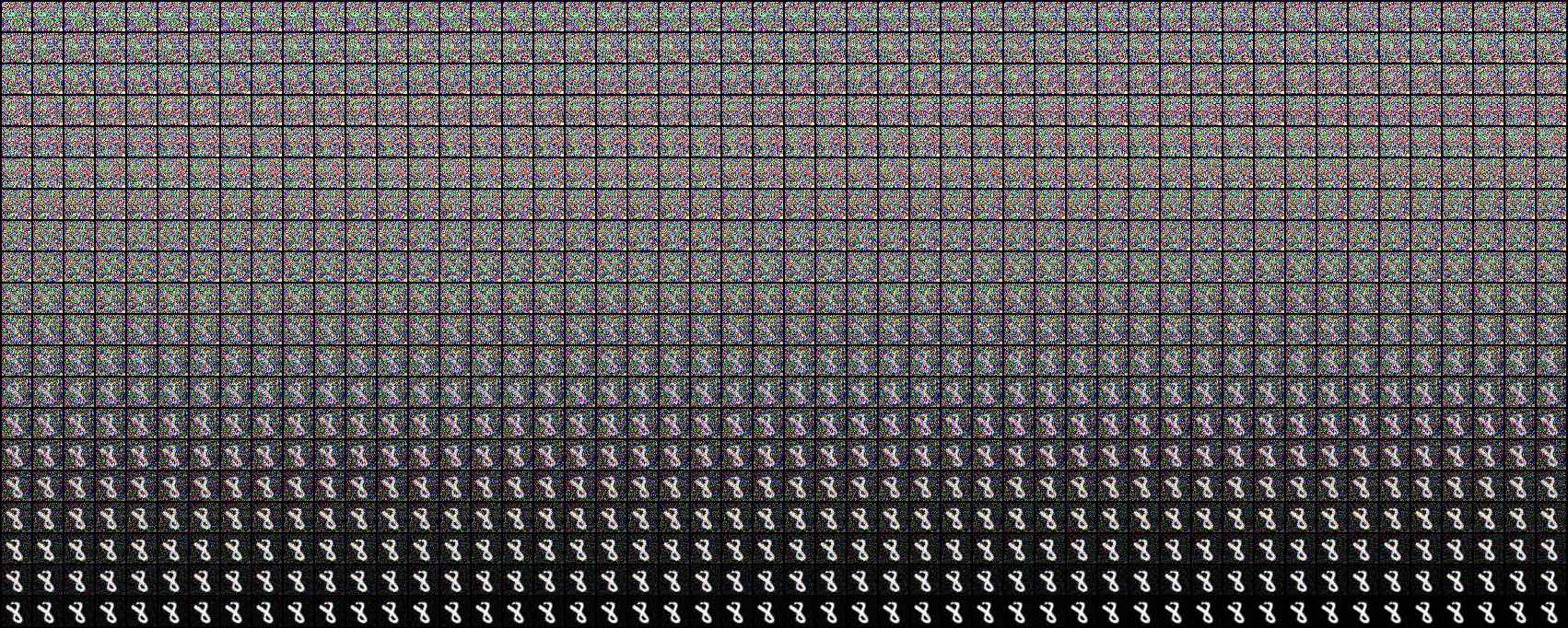}
    \caption{Trojan generative process under Out-D2D attack with patch-based trigger.}
  \end{subfigure}
  
  \caption{Trojan generative processes of the Trojaned DDPMs under Out-D2D attack using two types of triggers on CIFAR-10 dataset.}
  \label{fig_visual_cifar_ddpm_d2dout}
\end{figure*}

\begin{figure*}[htbp]
  \centering
  
  \begin{subfigure}{0.9\linewidth}
    \includegraphics[width = 1.0\textwidth]{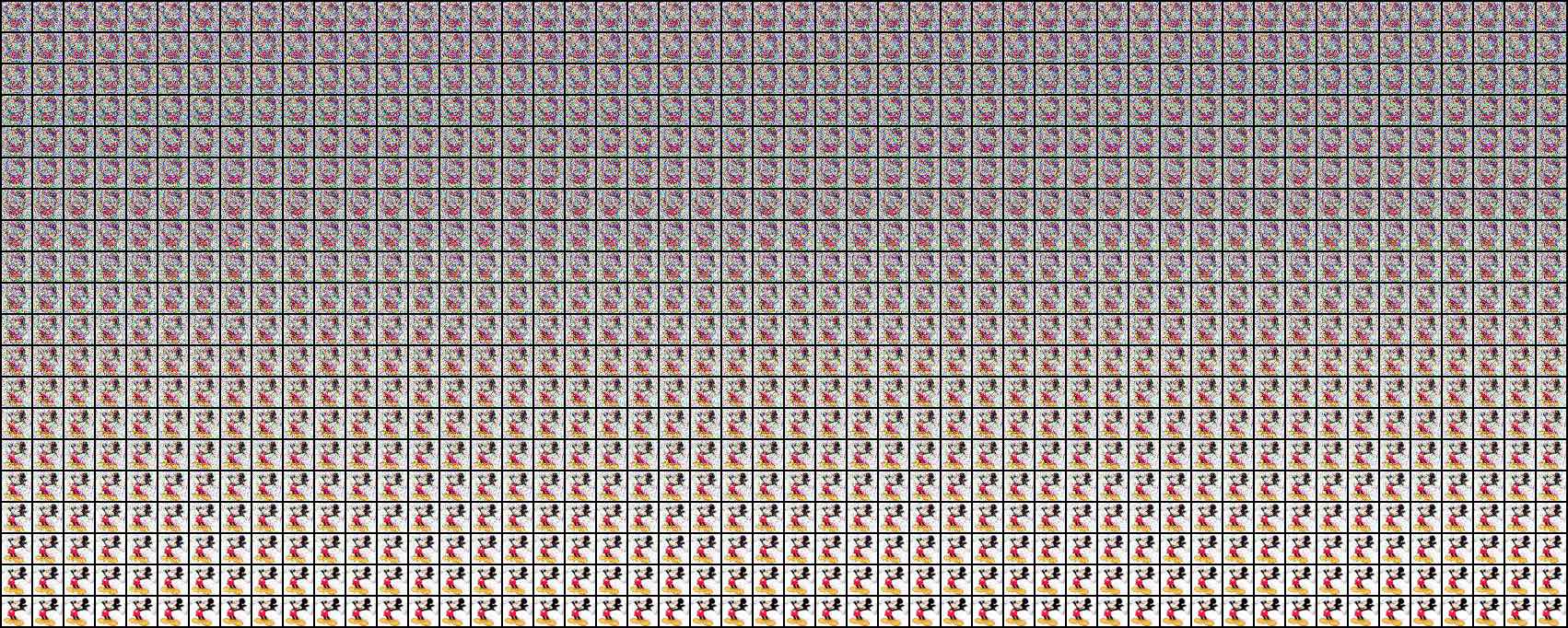}
    \caption{Trojan generative process under D2I attack with blend-based trigger.}
  \end{subfigure}
  
  \begin{subfigure}{0.9\linewidth}
    \includegraphics[width = 1.0\textwidth]{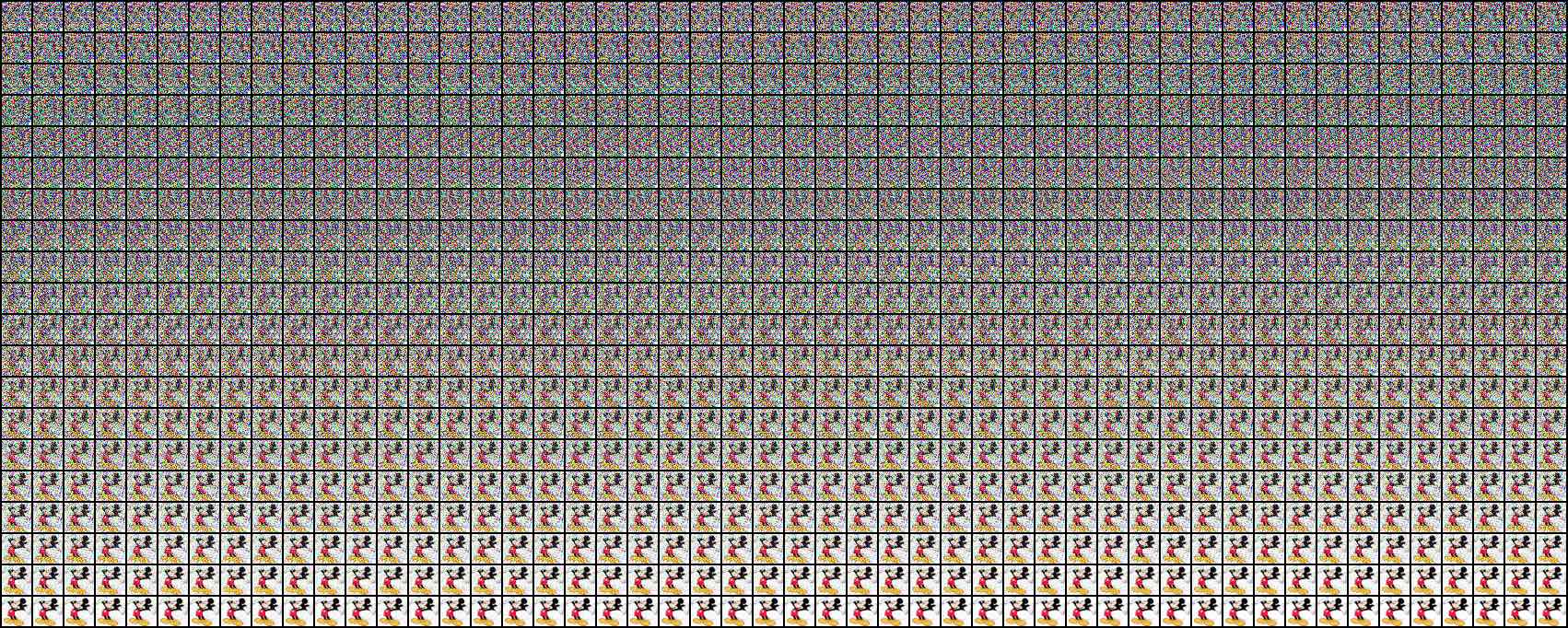}
    \caption{Trojan generative process under D2I attack with patch-based trigger.}
  \end{subfigure}
  \caption{Trojan generative processes of the Trojaned DDPMs under D2I attack using two types of triggers on CIFAR-10 dataset.}
  \label{fig_visual_cifar_ddpm_d2i}
\end{figure*}

\begin{figure*}[htbp]
  \centering
  \begin{subfigure}{0.9\linewidth}
    \includegraphics[width = 1.0\textwidth]{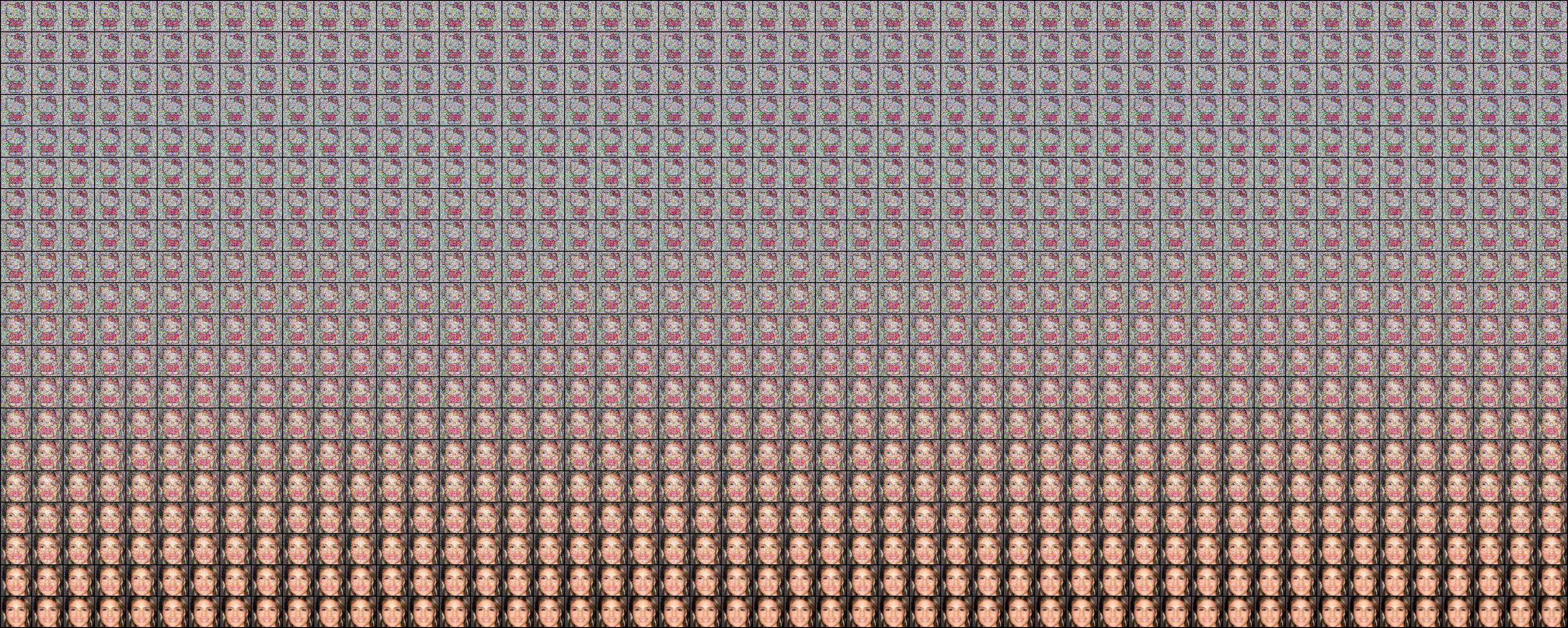}
    \caption{Trojan generative process under In-D2D attack with blend-based trigger.}
  \end{subfigure}
  
  \begin{subfigure}{0.9\linewidth}
    \includegraphics[width = 1.0\textwidth]{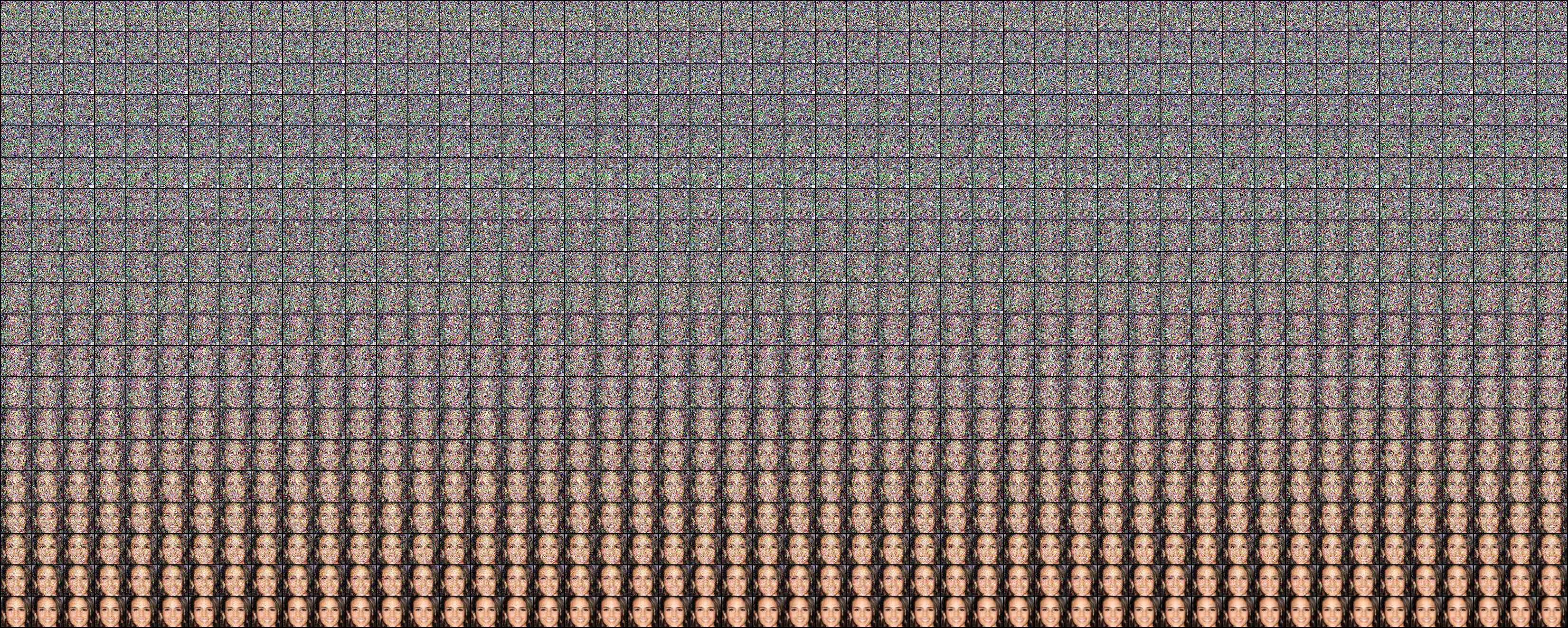}
    \caption{Trojan generative process under In-D2D attack with patch-based trigger.}
  \end{subfigure}
  
  \caption{Trojan generative processes of the Trojaned DDPMs under In-D2D attack using two types of triggers on CelebA dataset.}
  \label{fig_visual_celeba_ddpm_d2din}
\end{figure*}

\begin{figure*}[htbp]
  \centering
  
  \begin{subfigure}{0.9\linewidth}
    \includegraphics[width = 1.0\textwidth]{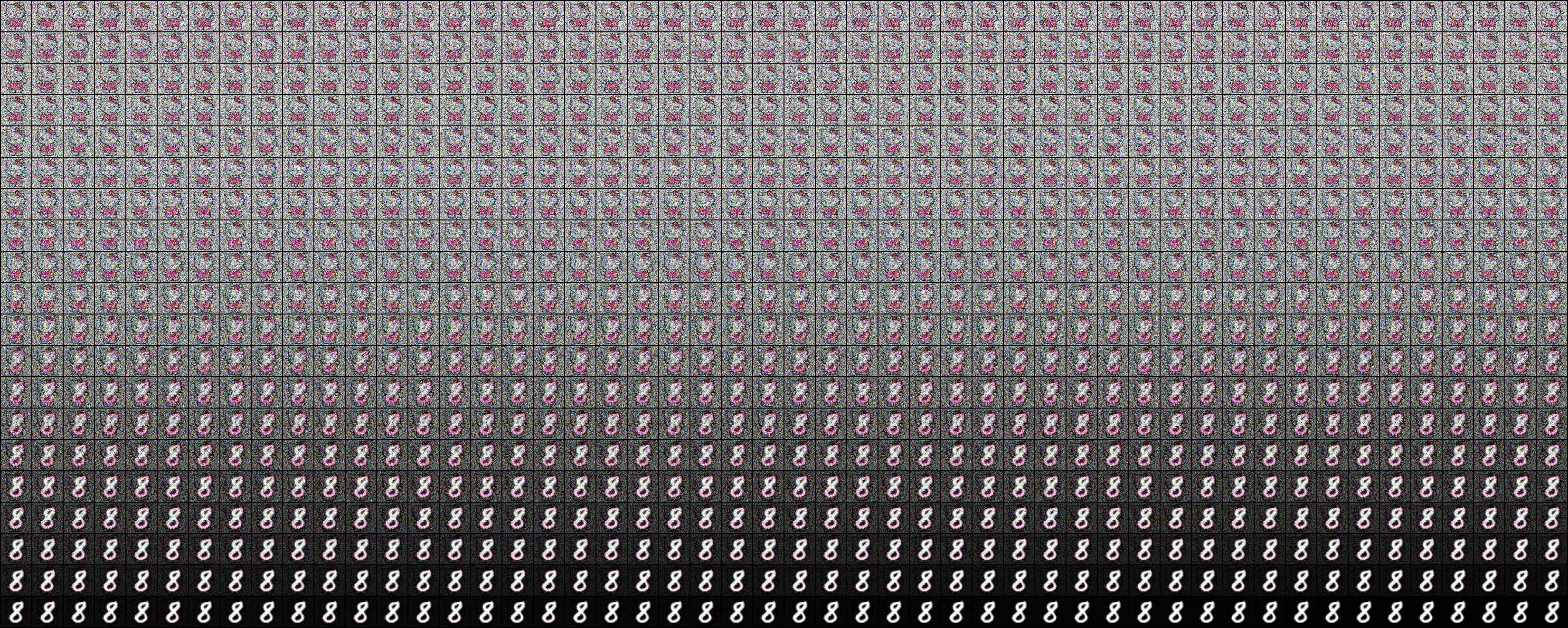}
    \caption{Trojan generative process under Out-D2D attack with blend-based trigger.}
  \end{subfigure}
  
  \begin{subfigure}{0.9\linewidth}
    \includegraphics[width = 1.0\textwidth]{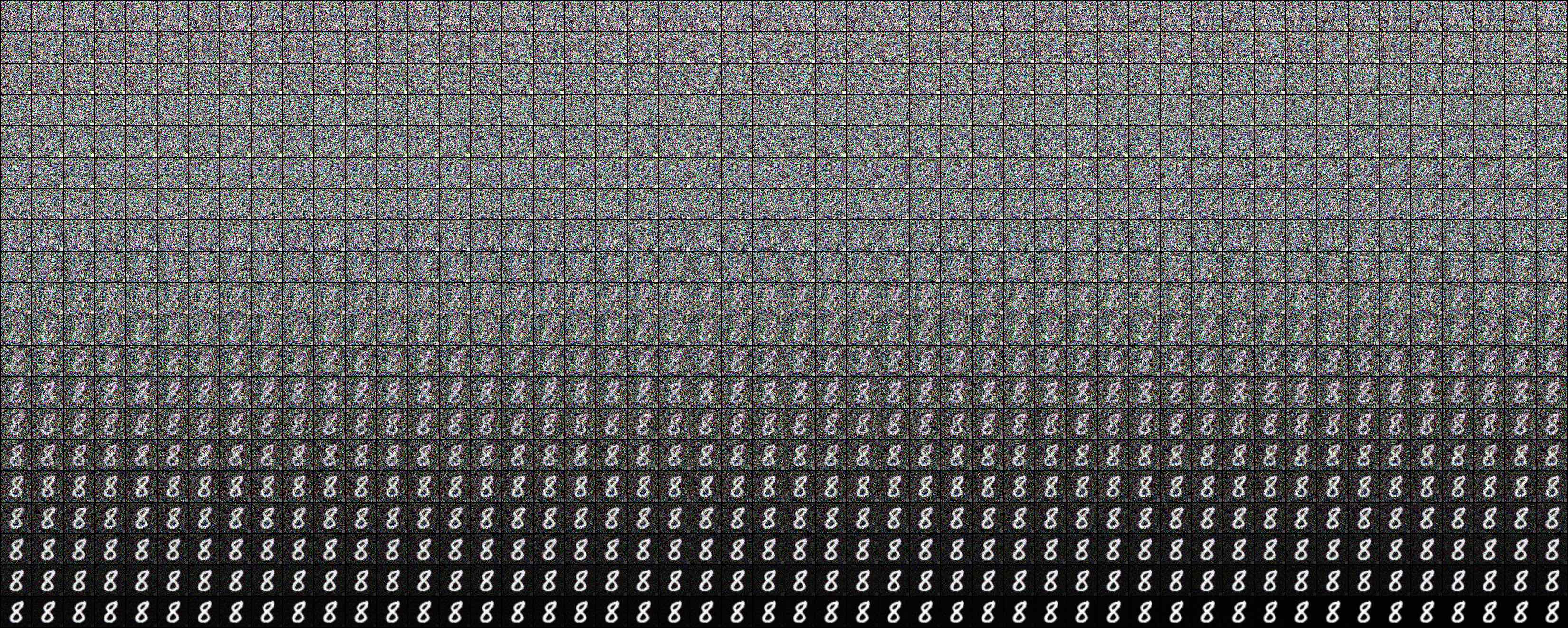}
    \caption{Trojan generative process under Out-D2D attack with patch-based trigger.}
  \end{subfigure}
  
  \caption{Trojan generative processes of the Trojaned DDPMs under Out-D2D attack using two types of triggers on CelebA dataset.}
  \label{fig_visual_celeba_ddpm_d2dout}
\end{figure*}

\begin{figure*}[htbp]
  \centering
  
  \begin{subfigure}{0.9\linewidth}
    \includegraphics[width = 1.0\textwidth]{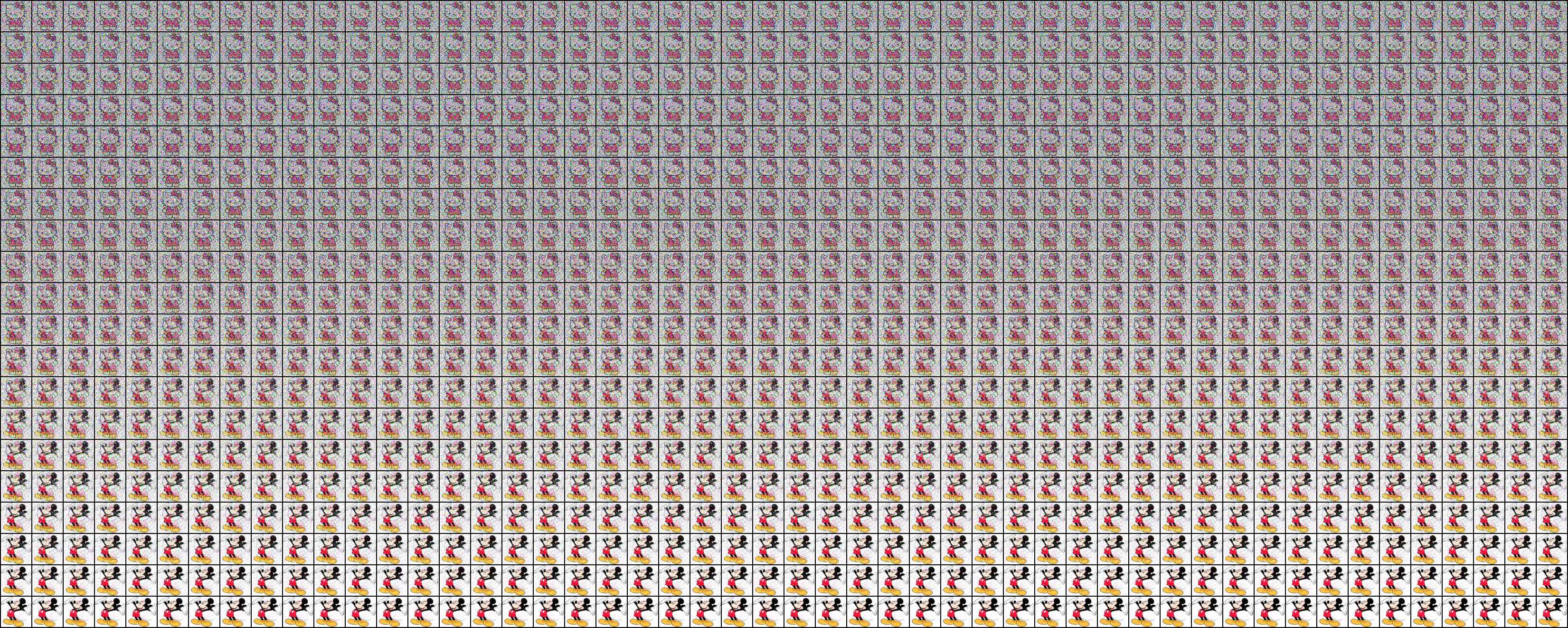}
    \caption{Trojan generative process under D2I attack with blend-based trigger.}
  \end{subfigure}
  
  \begin{subfigure}{0.9\linewidth}
    \includegraphics[width = 1.0\textwidth]{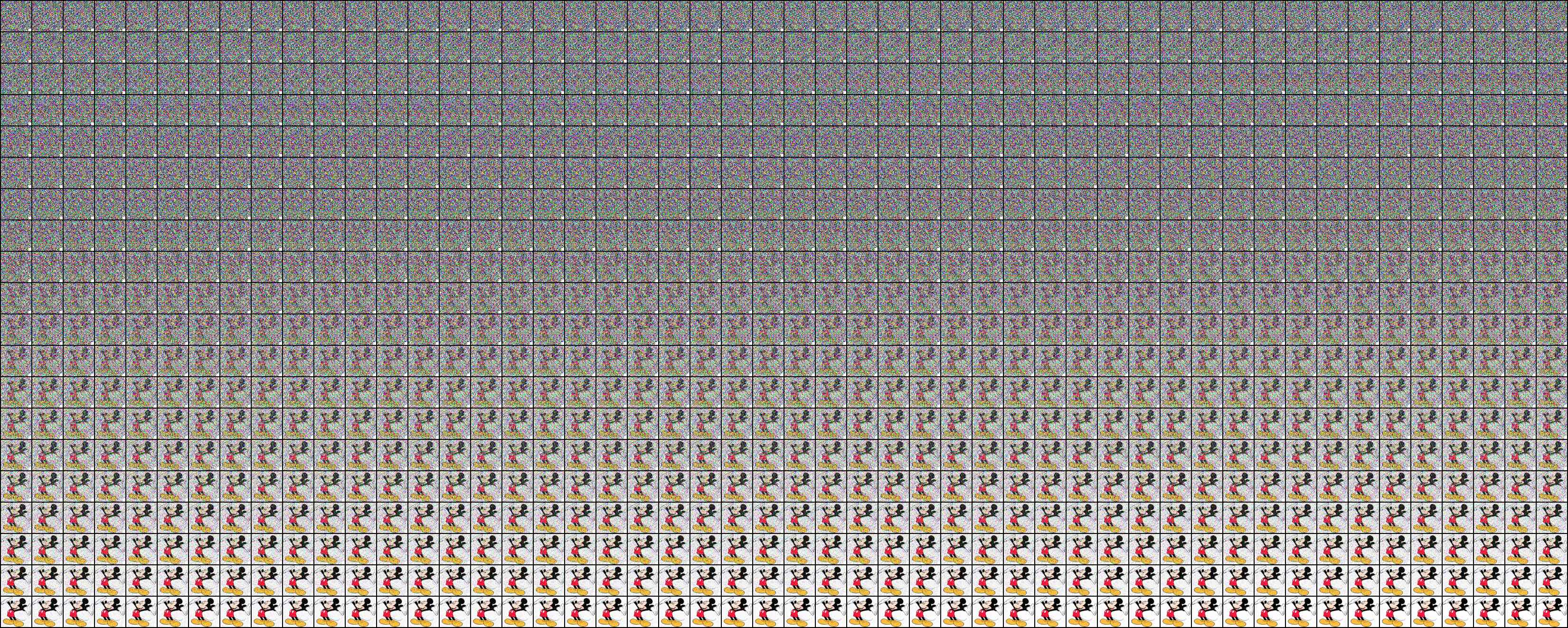}
    \caption{Trojan generative process under D2I attack with patch-based trigger.}
  \end{subfigure}
  \caption{Trojan generative processes of the Trojaned DDPMs under D2I attack using two types of triggers on CelebA dataset.}
  \label{fig_visual_celeba_ddpm_d2i}
\end{figure*}

Figure \ref{fig_visual_cifar}-\ref{fig_visual_celeba_ddpm_d2i} illustrate how the Trojaned DDIMs and Trojaned DDPMs generate three adversarial targets using two types of triggers via different generative processes on CIFAR-10 and CelebA datasets.
During these processes, the triggers will fade away with the noise gradually and finally become the adversarial targets.
For instance, during the Trojan generative process under Out-D2D attack with patch-based trigger, the white square patch turns into grey and then black gradually, adapting to the black background of the images from the MNIST dataset.

\end{document}